\documentclass[journal]{IEEEtran}
\usepackage{graphicx}
\usepackage{epsfig}
\usepackage{amsfonts}

\usepackage{amsmath}
\usepackage{hyperref}
\hypersetup{colorlinks=true, allcolors=blue, urlcolor=purple}
\usepackage{soulutf8}%
\usepackage[colorinlistoftodos]{todonotes}
\usepackage{color}
\usepackage{xargs}%
\usepackage{float}
\usepackage{wrapfig}
\usepackage{multirow}
\usepackage[symbol]{footmisc}
\usepackage{subcaption}
\usepackage{booktabs}
\usepackage{array}
\usepackage{bm}
\usepackage{algorithm}
\usepackage{listings}
\usepackage{xcolor}
\usepackage{bbding}
\usepackage{svg}

\usepackage{mathtools, stmaryrd}
\usepackage{xparse} \DeclarePairedDelimiterX{\Iintv}[1]{\llbracket}{\rrbracket}{\iintvargs{#1}}
\NewDocumentCommand{\iintvargs}{>{\SplitArgument{1}{,}}m}
{\iintvargsaux#1} %
\NewDocumentCommand{\iintvargsaux}{mm} {#1\mkern1.5mu..\mkern1.5mu#2}
\usepackage{stmaryrd}

\newcolumntype{?}{!{\vrule width 1pt}}
\usepackage{colortbl}
\definecolor{Gray}{gray}{0.85}
\newcolumntype{a}{>{\columncolor{Gray}}c}
\newcommand{\cut}[1]{}

\usepackage[switch]{lineno}
\begin{document}
\title{MVP: Meta Visual Prompt Tuning for Few-Shot Remote Sensing Image Scene Classification}

\author{Junjie Zhu, Yiying Li, Chunping Qiu, Ke Yang*, Naiyang Guan*, and~Xiaodong Yi
\IEEEcompsocitemizethanks{
\IEEEcompsocthanksitem Junjie Zhu, Ke Yang, Yiying Li, Chunping Qiu, Naiyang Guan, and Xiaodong Yi are with the National Innovation Institute of Defense Technology, Beijing, China (*Corresponding authors: Ke Yang, Naiyang Guan. E-mail: yangke13@nudt.edu.cn, nyguan@sina.com).
}
}
\markboth{SUBMIT TO IEEE TRANSACTIONS}%
{Shell \MakeLowercase{\textit{et al.}}: Bare Demo of IEEEtran.cls for Journals}
\maketitle

\begin{abstract} 
Vision Transformer (ViT) models have recently emerged as powerful and versatile models for various visual tasks. 
Recently, a work called PMF~\cite{hu2022pmf} has achieved promising results in few-shot image classification by utilizing pre-trained vision transformer models. However, PMF employs full fine-tuning for learning the downstream tasks, leading to significant overfitting and storage issues, especially in the remote sensing domain.
In order to tackle these issues, we turn to the recently proposed parameter-efficient tuning methods, such as VPT~\cite{jia2022vpt}, which updates only the newly added prompt parameters while keeping the pre-trained backbone frozen. Inspired by VPT, we propose the Meta Visual Prompt Tuning (MVP) method. Specifically, we integrate the VPT method into the meta-learning framework and tailor it to the remote sensing domain, resulting in an efficient framework for Few-Shot Remote Sensing Scene Classification (FS-RSSC).
Furthermore, we introduce a novel data augmentation strategy based on patch embedding recombination to enhance the representation and diversity of scenes for classification purposes.
Experiment results on the FS-RSSC benchmark demonstrate the superior performance of the proposed MVP over existing methods in various settings, such as various-way-various-shot, various-way-one-shot, and cross-domain adaptation. 
\end{abstract}


\IEEEpeerreviewmaketitle

\section{Introduction}
\label{sec:intro}
Few-shot remote sensing scene classification (FS-RSSC)~\cite{Zhenqi2022mknrs,zhu2017RSdeep} aims to classify remote sensing images into different categories using only a few labeled examples per category. This is a challenging but important machine learning task for practical applications such as land use classification and environmental monitoring, where obtaining large-scale and high-quality labeled datasets is expensive and time consuming. Transfer learning~\cite{pires2019RS_transfer}, meta-learning~\cite{li2021meta_rs}, and metric learning~\cite{Zhenqi2022mknrs} have been employed for FS-RSSC tasks. These methods primarily use convolutional neural networks (CNNs) that are typically restricted to smaller models with fewer parameters, such as Conv4~\cite{vinyals2016matching}, ResNet12~\cite{he2016resnet}, and ResNet18~\cite{he2016resnet}.

\begin{figure} [t] \centering
\centering
\includegraphics[scale=0.25]{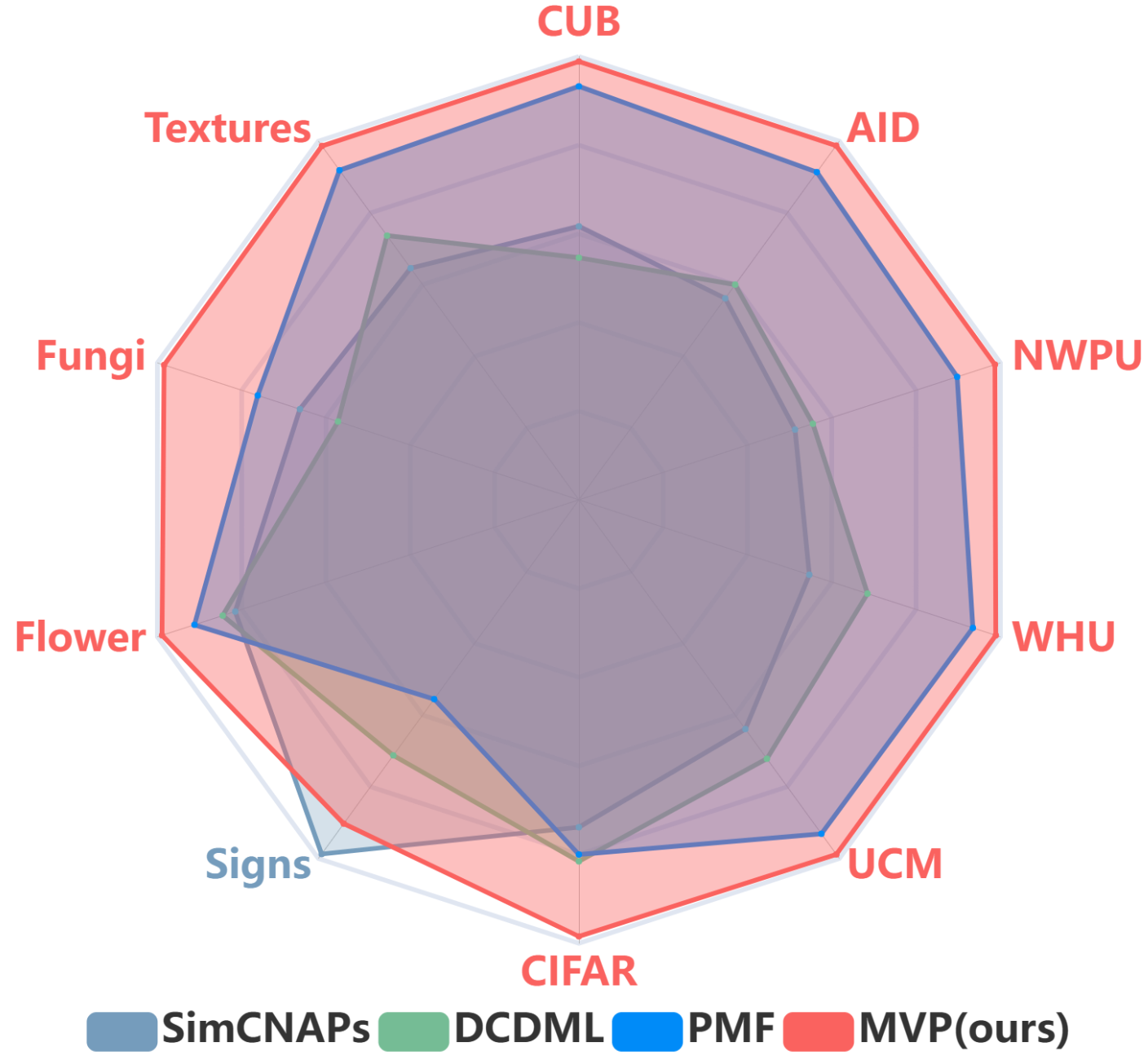}
\caption{MVP \cut{versus}vs. SoTA FS-RSSC algorithms (i.e., SimCNAPs~\cite{simcnps}, DC-DML~\cite{li2022rsAIFS}, PMF~\cite{hu2022pmf}): A succinct assessment of their accuracy performance on AIFS dataset~\cite{li2022rsAIFS}.}  
\label{fig:acc_overview}
\end{figure}
Recently, Vision Transformers (ViT) have yielded remarkable achievements in various visual tasks. The prospect of applying them to the field of few-shot learning is highly attractive and holds significant appeal.
For instance, PMF~\cite{hu2022pmf}, a ViT-based method consisting of a three-stage learning pipeline, has made significant progress in few-shot classification tasks.
PMF pre-trained the ViT model on unsupervised external data, then meta-trained the model on base categories, and finally fine-tuned the model on a novel task. They showed that this simple transformer-based pipeline yields surprisingly good performance on standard benchmarks such as Mini-ImageNet~\cite{vinyals2016matching}, CIFAR-FS~\cite{bertinetto2018meta}, CDFSL~\cite{guo2020broader_metatuning}, and Meta-Dataset~\cite{triantafillou2019metadataset}.

During the three stages of learning, PMF utilizes full fine-tuning to update network weights. However, this brings about two significant issues, especially in remote sensing applications. Firstly, compared to the few-shot classification task in natural images, the remote sensing domain faces a more severe issue of sample scarcity, and thus fully fine-tuning ViT with a large number of weights can lead to severe overfitting problems. Second, training a ViT model for each remote sensing task is unfeasible due to storage limitations on the satellite or drone platforms where the algorithm is deployed. Our experiment results also indicate that ViT models based on a full fine-tuning strategy exhibit lower efficacy in solving FS-RSSC tasks.

\begin{figure*}[ht]
\centering
\begin{minipage}[t]{0.85\linewidth}
\epsfig{figure=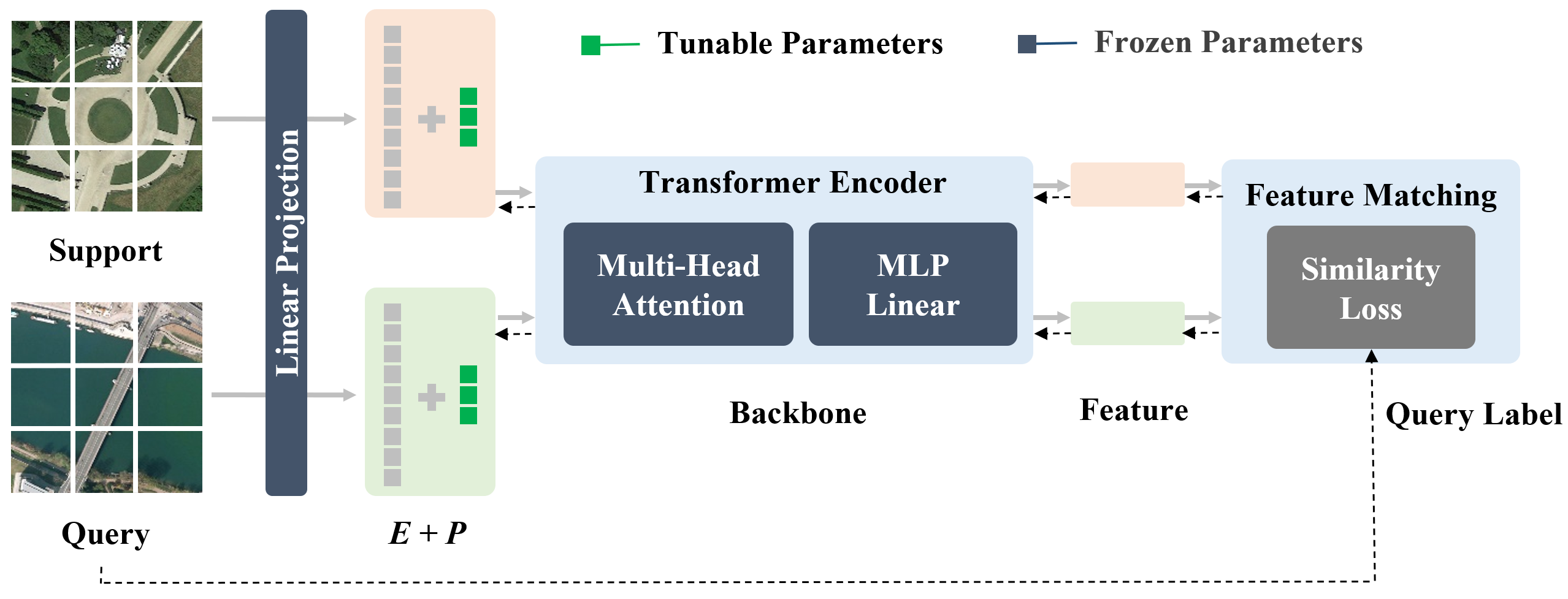,width=\linewidth}
\end{minipage}
	\caption{Meta Prompt Tuning framework for Remote Sensing Scene Classification. Query in the dashed box: actual image in meta-training and pseudo-query generated via a novel data augmentation method in meta fine-tuning. ‘E’ represents original backbone embedding tokens; ‘P’ represents newly added prompt tokens.} 
\label{fig:arch}
\end{figure*} 
In order to address these issues, we turn to explore a fine-tuning method for ViT models that is suitable for FS-RSSC tasks. A potential solution is the Parameter-Efficient Fine-Tuning (PEFT)~\cite{brown2020language_prompt} methods, which have received considerable attention in natural image recognition recently. In the PEFT paradigm, only a small number of newly added parameters are updated during training, while the pre-trained backbone is kept frozen. For instance, Visual Prompt Tuning (VPT)~\cite{jia2022vpt} is a recently proposed visual PEFT method that adds prompt tokens to the input space and only updates the newly added parameters. The tuned parameters for each downstream task are less than 1\% of model parameters, thus it can reduce the storage demand and effectively alleviate the model overfitting issues of remote sensing applications.

Taking inspiration from VPT, we propose the Meta Visual Prompt Tuning (MVP) method as an effective approach to address FS-RSSC tasks. Within the framework of meta-learning, MVP leverages prompt tuning to adapt a pre-trained ViT model to new tasks with limited data and computational resources. Unlike previous methods that fine-tune the entire ViT model, MVP only updates the newly added prompt parameters while keeping the pre-trained ViT backbone networks fixed. Specifically, MVP embeds the prompt parameters into a novel parameter-efficient meta-learning framework. In the meta-training phase, MVP learns to learn a good initialization for the newly added prompt parameters on multiple sets of FS-RSSC source tasks. We denote the optimal initialization prompt parameters by $\bm{\theta}$. In the meta fine-tuning phase, MVP fine-tunes $\bm{\theta}$ with a few gradient steps on the target task and then makes predictions for remote sensing scene categories.

Additionally, we design a novel data augmentation method for meta fine-tuning. Our method is motivated by the observation that remote sensing scene images of the same category tend to have high consistency, which may lead to model overfitting and poor generalization to variations in imaging conditions. To address this issue, we propose to enhance the diversity of remote sensing scenes by embedding image patches of other categories into the current image, inspired by image patch recombination research~\cite{gong2022twopath_RS}. Specifically, a new image patch recombination method based on the ViT network is designed, which operates on image patch embeddings after the linear projection of ViT. Moreover, we randomly select and swap some patch embeddings of an input image with those from other images in the same batch. Experiments show that our data augmentation method can effectively enhance the generalization performance to new categories in the meta fine-tuning stage.

Real-world FS-RSSC tasks exhibit two key characteristics: the number of categories and samples in new tasks varies, and the data distribution is unpredictable, thus necessitating cross-domain adaptation~\cite{li2022rsAIFS}. 
The AIFS dataset~\cite{li2022rsAIFS} meets both criteria as the benchmark for evaluation.
We thoroughly evaluated our proposed MVP model on the AIFS dataset through comprehensive experiments, under various-way-various-shot, various-way-one-shot, and cross-domain adaptation settings.
Fig.~\ref{fig:acc_overview} shows an overview of the comparison of the results and Fig.~\ref{fig:arch} depicts the pipeline of the MVP model. Our MVP demonstrated superior performance on various challenging in-domain and cross-domain benchmarks, significantly surpassing the existing methods.
Our contributions can be summarized as follows:
\begin{itemize}
\setlength{\parsep}{0pt} 
\setlength{\topsep}{0pt} 
\setlength{\itemsep}{0pt}
\setlength{\parsep}{0pt}
\setlength{\parskip}{0pt}
    \item To the best of our knowledge, our proposed MVP is the first study to explore Parameter-Efficient Tuning (PETuning) on remote sensing applications.
    \item We integrate PETuning into the meta-learning framework by employing the meta-learning paradigm to initialize newly added prompt parameters, facilitating rapid adaptation to new FS-RSSC tasks. 
    \item Our proposed MVP empowers large transformer models to perform well in situations where there is only limited data available, significantly alleviating the overfitting issue.
    \item We propose a data augmentation method tailored explicitly for ViT-based FS-RSSC models to enhance their adaptability to remote sensing scenes.
    \item Our MVP demonstrates exceptional performance on the challenging FS-RSSC dataset.
\end{itemize}

\section{Related Work}
\subsection{Scene Classification With Few-Shot Learning} 
Few-shot remote sensing scene classification (FS-RSSC) is a challenging task due to the limited availability of annotated data for model training. Classical methods can be roughly divided into two categories: metric-based and meta-learning-based~\cite{cheng2021spnet}. Metric-based methods learn a feature space or distance function to measure the similarity between classes~\cite{Zhenqi2022mknrs,li2020rs_metric,li2020dla_rs}. For instance, RS-MetaNet~\cite{li2020rs_metric} introduces a novel balance loss to provide better linear segmentation planes for scenes in different categories. Another example is DLA-MatchNet~\cite{li2020dla_rs}, which proposes an approach to automatically discover discriminative regions. MCMNet~\cite{zeng2022clRS} takes this a step further by proposing a multi-scale covariance network to optimize the manifold space.

On the other hand, meta-learning-based methods~\cite{finn2017model, zhang2021metaRS} aim to learn a meta-model that can quickly adapt to new tasks with few gradient updates. For instance, MetaRS~\cite{zhang2021metaRS} explores the use of meta-learning to improve the generalization capability of deep neural networks (DNN) on remote sensing scene classification with limited training data. Another example is PTMeta~\cite{ma2021rs_ptmeta}, which applies parameter transfer to fix the parameters in a DNN to relax the problem of training a large number of parameters within a meta-learning framework.

A review of these methods reveals that most employ shallow CNNs as their backbone network. While this approach can mitigate overfitting when training data is limited, it also constrains further improvements in classification performance. Recently, Vision Transformers (ViT) have demonstrated promising results in visual tasks~\cite{dosovitskiy2020vit,hu2022pmf}, and there have been efforts to apply ViT-based approaches to large-scale remote sensing classification~\cite{roy2022rsvit_large,bi2022rsvit_large}. However, research on the application of ViT-based methods to FS-RSSC tasks remains limited.

\subsection{Efficient Tuning for Visual Transformer} 
Efficient tuning of pre-trained language models (PLMs) has attracted much attention recently~\cite{devlin2018bert_nlp, logan2021cutting_nlp}, as it aims to reduce the parameter size and computational cost of fine-tuning PLMs on downstream tasks. Various methods have been proposed to achieve efficient tuning, such as adding light-weight adapter modules~\cite{rebuffi2018efficient_adapter, zhang2020side_adapter}, inserting task-specific prefix tokens~\cite{li2021prefix}, or appending prompt tokens~\cite{lester2021power_prompt} to the PLMs and only updating these additional parameters while keeping the PLMs frozen. Prompt tuning has recently been extended to visual tasks. For instance, Visual Prompt Tuning (VPT)~\cite{jia2022vpt} achieves higher accuracy than full fine-tuning by updating only 1\% of the model’s parameters. However, prompt tuning also faces some challenges in few-shot learning settings, where each task has only a small amount of labeled data available. There is still a lack of sufficient research on how to leverage prompt tuning for few-shot learning scenarios. In this article, we inherit the VPT technique and propose a novel method that advances the state-of-the-art performance on FS-RSSC tasks.

\subsection{Data Augmentation} 
In both general and few-shot image classification, data augmentation expands the number of available images per class and generates novel classes and tasks~\cite{shorten2019datasurvey}. Techniques range from simple rotations~\cite{gidaris2018rotate} and crops~\cite{zhang2020data_crop} to more refined strategies such as cutmix~\cite{yun2019cutmix} and mixup~\cite{kim2020data_mixup}. Some methods use GAN networks to emulate the target data distribution~\cite{li2020data_gan}. Recent research has shown that data augmentation has different effects on the meta-training and meta-testing stages of the meta-learning pipeline~\cite{ni2021metadata}. For instance, increasing the number of query samples and tasks during meta-testing improves the performance of meta-learners more than increasing the number of support samples during meta-training. Compared to general few-shot datasets, FS-RSSC datasets have a smaller volume~\cite{Zhenqi2022mknrs}. To overcome this data scarcity problem, several methods have been proposed to augment the training data for FS-RSSC in different ways. For example, the quad-patch method~\cite{gong2022twopath_RS} generates synthetic samples by cutting and reassembling patches from existing images, while the spatial vector enhancement method~\cite{yang2021data_free} simulates the distribution of neighboring classes to enrich the feature space. However, these methods do not consider the specific properties of ViT as the backbone network. In this article, we present a novel data augmentation method that is customized for the structural features of ViT and the attributes of remote sensing images and fully unleashes the potential of the ViT architecture.

\section{METHODOLOGY}  
\subsection{Overview}  
\textbf{Problem Definition.}
Few-shot remote sensing scene classification (FS-RSSC) is a task that requires a model to quickly and accurately classify unseen scene images with only a few annotated samples~\cite{Zhenqi2022mknrs}. This task is motivated by the challenge of domain adaptation in remote sensing images, which are often collected from different sensors, regions, and seasons, resulting in a large domain gap between the source and target domains. Moreover, the number of annotated samples varies significantly across different tasks, making it necessary to train models that can cope with different numbers of annotated samples. Formally, the annotated dataset, referred to as the support set $S$, consists of $C$ categories (way) with $K$ samples (shot) per category, where $C \in [5, \text{MAXWAY}]$ and $K \in [1, \text{MAXSHOT}]$. The model is trained on the support set and then used to predict the categories of the query set $Q$, which contains unlabeled images from the same categories as the support set.

\textbf{Meta-learning Process.}
Meta-learning methods have shown promise for the FS-RSSC task. The basic meta-learning process consists of two stages: meta-training and meta fine-tuning \cite{li2021meta_rs, zhang2021metaRS}. With the development of ViT, the PMF~\cite{hu2022pmf} method enhances the meta-learning process by adding a pre-training phase. PMF proposes a new three-stage pipeline: pre-training, meta-training, and meta fine-tuning. The backbone network is first pre-trained on a large-scale external dataset such as ImageNet~\cite{deng2009imagenet}, then meta-trained on multiple source datasets, and finally, meta fine-tuned on a target dataset with limited annotated support samples. Note that the source and target datasets have non-overlapping domains.

\textbf{Prompt-based Meta-learning.}
During the three stages of learning, PMF utilizes full fine-tuning to update the backbone network. However, this brings about two significant challenges, especially in remote sensing applications. One challenge is how to avoid overfitting when fine-tuning the whole ViT model on limited support samples. Another challenge is how to reduce the storage space required to store different ViT models for different tasks. To address these issues, we propose a meta-visual prompt tuning (MVP) framework. Our research aims to efficiently fine-tune the pre-trained ViT backbone models for the FS-RSSC task.

\begin{figure}[t]
\centering
\begin{minipage}[t]{0.92\linewidth}
\epsfig{figure=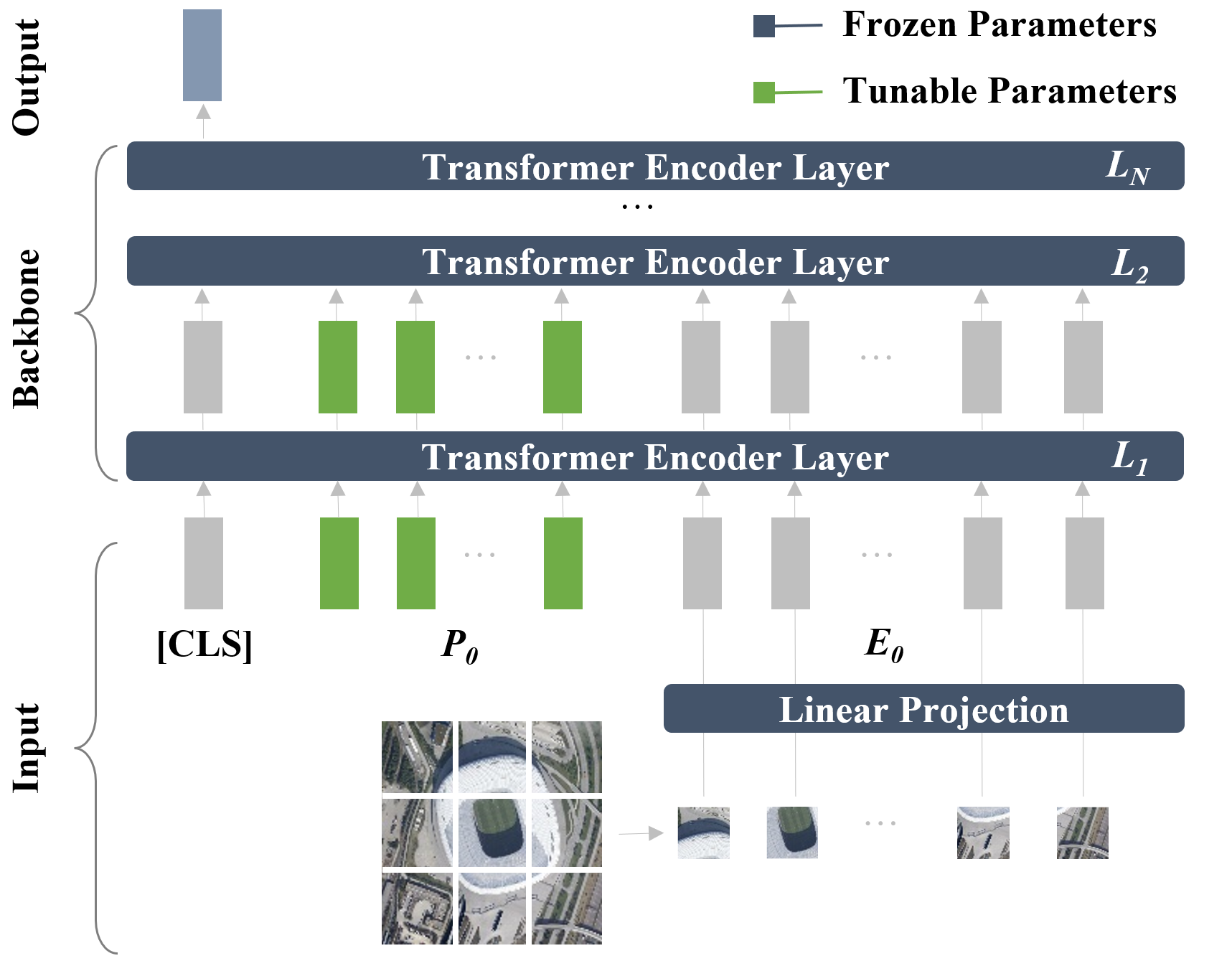,width=\linewidth}
\end{minipage}
	\caption{The network structure of the proposed Meta Visual Prompt Tuning (MVP) model.} 
\label{fig:model}
\end{figure}
Given a pre-trained ViT backbone network, MVP introduces a new prompt module into the input space of the pre-trained ViT model (as shown in Fig.~\ref{fig:model}). The prompt parameters do not need to undergo the pre-training stage as in PMF. During the meta-training and meta fine-tuning stages, MVP only updates the prompt parameters to fit the FS-RSSC task features, while the whole ViT network is frozen. 

In the meta-training stage of MVP, we optimize the prompt parameters on multiple source datasets that are domain-disjoint from the target dataset. We use a meta-learning algorithm that mimics the FS-RSSC task by sampling episodes from the source datasets. An episode is a few-shot learning task that contains a support set and a query set with the same categories but different images. We use a prototypical loss function~\cite{snell2017prototypical} to evaluate the classification performance of the MVP model on the query set and update the prompt parameters using gradient descent.

After initializing the prompt parameters in meta-training, the MVP model is able to adapt to the target dataset of remote sensing scenes in the meta fine-tuning phase. These target scenes are completely new and different from the source dataset. The MVP model performs meta fine-tuning on auxiliary tasks that are based on support data, using a novel data augmentation method. After meta fine-tuning, the MVP model can classify all the remaining unlabeled query data.

\subsection{Model Architecture}  
\textbf{ViT Backbone Networks.}
The standard ViT~\cite{dosovitskiy2020vit} is employed as our backbone network to address the FS-RSSC task. As input to the ViT backbone network, the image $x \in \mathbb{R}^{3 \times H \times W}$ is initially partitioned into $m$ fixed-size patches $\{I_j \in \mathbb{R}^{3 \times h \times w} \mid j \in \mathbb{N}, 1 \le j \le m \}$. Subsequently, each patch is projected into $d$-dimensional feature embedding with positional encoding~\cite{dosovitskiy2020vit}:
\begin{align}
\label{eq:patch}
\bm{e}_0^j = \texttt{Embed}(I_j), 
\end{align}
where $\bm{e}_0^j \in \mathbb{R}^{d}$. Following this, the collection of image patch embeddings  
$\bm{E}_{i}=\{\bm{e}_i^j \in \mathbb{R}^d \mid i \in \mathbb{N}, 1 \le i \le N \}$
is employed as the inputs to the ($i$+$1$)-th Transformer layer $L_{i+1}$. Formally, the entire ViT backbone networks can be articulated as:
\begin{align} 
\label{eq:vit}
    [\bm{CLS}_i, \bm{E}_i] &= L_i([\bm{CLS}_{i-1}, \bm{E}_{i-1}]), \\
    f_{\theta}(x) &= \bm{CLS}_N,
\end{align}
where $\bm{CLS}_{i-1} \in \mathbb{R}^{d}$ refers to the class token in the input sequence of $L_i$. Furthermore, $\bm{CLS}_N$ in the output of the final layer is employed as the feature representation $f(x)$ of the input image $x$. Moreover, $\theta$ represents the model parameters of ViT.

\textbf{Prompt-based ViT Networks.}
In line with the VPT~\cite{jia2022vpt} model and given a pre-trained ViT backbone network, a set of prompt tokens 
$\bm{P}_{i}=\{\bm{p}_i^t \in \mathbb{R}^d \mid t \in \mathbb{N}, 1 \le t \le p \}$
is concatenated into the input space of the transformer layer. Formally, the ViT architecture in Eq.~\ref{eq:vit} can be replaced by:    

\begin{align}   
\label{eq:vpt}
    [\bm{CLS}_i, \bm{P}_i, \bm{E}_i] &= L_i([\bm{CLS}_{i-1}, \bm{P}_{i-1}, \bm{E}_{i-1}]),  \\
    f_{{\theta}^{'}}(x) &= \bm{CLS}_N,
\end{align}
where $[\bm{x}_{i-1}, \bm{P}_{i-1}, \bm{E}_{i-1}] \in \mathbb{R}^{(1+p+m)\times d}$. ${\theta}^{'}$ denotes the model parameters of ViT, as well as the additional prompt parameters ${\theta}^{P}$. The network structure of our proposed model, Meta Visual Prompt Tuning (MVP), is illustrated in Fig.~\ref{fig:model}. 

Throughout the meta-training and meta fine-tuning phase, only the newly added prompt parameters ${\theta}^{P}$ are updated, while all other parameters of the ViT backbone network remain unchanged. Therefore, a classification task involving the prediction of label $y$ can be represented as follows:

\begin{align}   
\label{eq:vpt}
&\arg\max_{{\theta}^{P}} \sum_{\bm{x}} \log p\big(y|f_{{\theta}^{'}}(\bm{x});{\theta}^{P}\big). 
\end{align}

\begin{figure}[t!]
\centering
\begin{minipage}[t]{0.7\linewidth}
\epsfig{figure=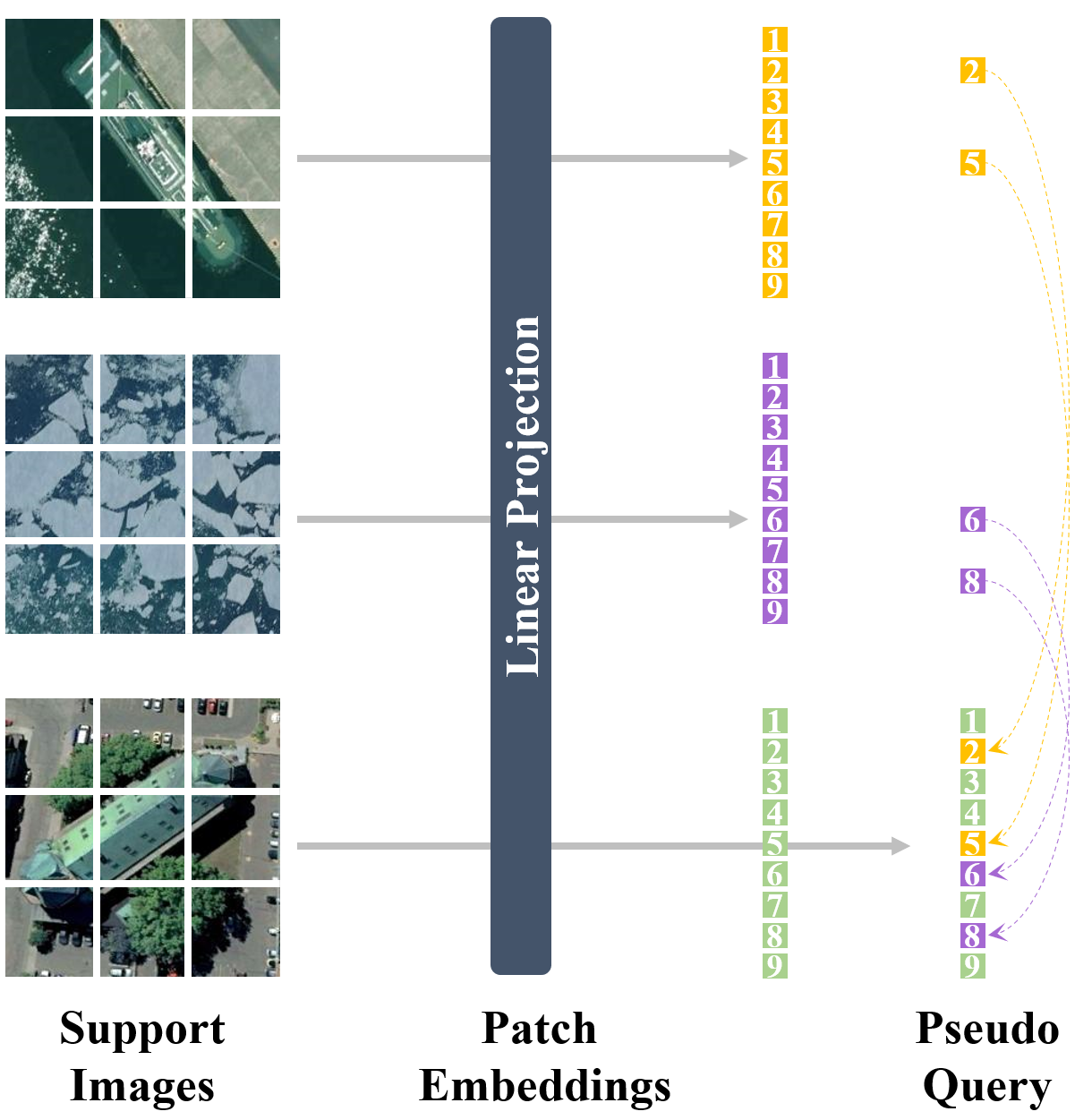,width=\linewidth}
\end{minipage}
	\caption{Random Patch Recombination: a novel data augmentation method proposed in this article.} 
\label{fig:data_process}
\end{figure}

\subsection{Meta Fine-Tuning Process} \label{sec:dataaug}
Meta fine-tuning for target few-shot tasks requires effectively using a small amount of labeled support data and achieving meta fine-tuning in a few steps. A common solution is to use data augmentation to expand the support set~\cite{hu2022pmf}. For a few-shot task $T=\{S, Q\}$, where $S$ is a set of labeled support images, and $Q$ are other unlabeled query images, one method is to create an auxiliary task $T'=\{S, Q'\}$, where the pseudo-query $Q’=\text{augment}(S)$ is composed of augmented support images. This auxiliary task-based meta fine-tuning process enhances the model's adaptability to novel tasks by leveraging augmented data.

In this work, we propose a novel data augmentation method called Random Patch Recombination (RPR), which is designed specifically for ViT-based models.
As shown in Fig.~\ref{fig:data_process}, the RPR method is applied after the linear projection of the data. Unlike traditional methods~\cite{hu2022pmf, ni2021metadata} that augment the data before feeding it into the backbone networks, our method can fully utilize the structure of ViT.
Specifically, for the patch embeddings $\bm{E}=\{\bm{e}^{pos} \in \mathbb{R}^d \mid pos \in \mathbb{N}, 1 \le pos \le m \}$ of a given image  in a support set $S$, we select a subset of $\bm{E}$ and denote their corresponding positions as $\{\bm{pos} \in \mathbb{R}^{m'}, 1 \leq m'\leq m \}$, with a recombination rate of $\alpha$. Then, we replace the selected patches with patches from other images in the support set $S$ that have the same positions $\bm{pos}$. Algorithm~\ref{alg:alg} summarizes the process in detail.

\subsection{Loss Functions}  
With the prompt-based ViT backbone networks, we discuss how to define our training objective. We denote the feature representation of support and query image as $f_{{\theta}^{'}}(x^s)$ and $f_{{\theta}^{'}}(x^q)$ and write them as $f_{{\theta}^{'}}^s$ and $f_{{\theta}^{'}}^q$, for simplicity. Following the prototypical networks~\cite{snell2017prototypical}, the prototype vectors of support data are denoted as $\bm{\Omega	}=\{\bm{\mu}_c \in \mathbb{R}^d \mid c \in \mathbb{N}, 1 \le c \le C \}$. Where $\bm{\mu}_c = \frac{1}{\left| S_c \right|} \sum_{i: y^s_i = c} f_{{\theta}^{'}}^{s_i}$ is the prototype of class $c$ and ${\left| S_c \right|} = \sum_{i: y^s_i = c} 1$. Then the probability of a query image $x^q$ is defined as a function of its similarity to the prototypes of support data:        
\begin{align}
\label{eq:proto}
p(y^q=c | x^q) = \frac{\exp\big( -d(f_{{\theta}^{'}}^q, \bm{\mu}_c) \big) }{\sum_{c'} \exp\big( -d(f_{{\theta}^{'}}^q, \bm{\mu}_{c'}) \big)},
\end{align}
where $d$ is cosine distance. Finally, the training objective $\mathcal{L}$ of our proposed MVP is to minimize the negative log-likelihood and $\mathcal{L}=-\log p({y^q}=c|{x^q})$, which can be further written as:      

\begin{align}
\label{equ:loss}
\mathcal{L} = \frac{1}{\left| S_c \right|} \left[ d(f_{{\theta}^{'}}^q,{\bm{\mu}}_c)+\log \sum\limits_{c'}{\exp -d(f_{{\theta}^{'}}^q,{\bm{\mu}}_{c'})} \right].  
\end{align}

\begin{algorithm}[t]
   \caption{PyTorch pseudo code for data augmentation}
   \label{alg:alg}
    \definecolor{codeblue}{rgb}{0.25,0.5,0.5}
    \lstset{
      basicstyle=\fontsize{7.2pt}{7.2pt}\ttfamily\bfseries,
      commentstyle=\fontsize{7.2pt}{7.2pt}\color{codeblue},
      keywordstyle=\fontsize{7.2pt}{7.2pt},
    }
\begin{lstlisting}[language=python]
# Inputs: Patch embeddings of images in a support set. 
# Outputs: Recombination of input patch embeddings.

def random_recombine(emb_x, rate):
    # x: [bs, pos, dim]
    # rate: selection rate of \alpha
    pos = emb_x.size(1) 
    rate = int(pos * rate)
    # iterate over all image patch embeddings 
    for emb_i, _ in enumerate(emb_x):  
        select_pos = random.sample(range(0, pos), rate)
        emb_x = replace(emb_x, emb_i, select_pos)
        
def replace(emb_x, emb_i, select_pos):
    # iterate over all selected positions
    bs = emb_x.size(0)
    for pos_i in select_pos:
        # note: probability of p(i=j) is approximately 0
        emb_j = random.randint(0, bs-1)
        emb_x[emb_i, pos_i, :] = emb_x[emb_j, pos_i, :] 
    return emb_x
\end{lstlisting}
\end{algorithm}

\section{EXPERIMENTS}
In this section, we first introduce the evaluation dataset and the implementation details briefly but comprehensively. Then, we present ablation experiments that demonstrate the significant effectiveness of the MVP design. Finally, we contrast the proposed MVP with state-of-the-art (SoTA) peer competitors.

\subsection{Experimental Setup}
\label{experimental_setup}
 
\textbf{Datasets.}
We evaluate our methods and SoTA algorithms using a challenging FS-RSSC benchmark named AIFS-DATASET~\cite{li2022rsAIFS}. This benchmark is a variant of the META-DATASET~\cite{triantafillou2019metadataset} that includes a collection of remote sensing datasets. The AIFS-DATASET is composed of two subsets: in-domain and out-of-domain sets. The in-domain set consists of six open-source datasets that do not feature remote sensing scenarios: CUB-200-2011, Describable Textures, Fungi, VGG Flower, Traffic Signs, and CIFAR100. These six datasets are partitioned such that approximately 70\%, 15\%, and 15\% of data are assigned to training, validation, and testing sets, respectively. Out-of-domain set comprises four remote-sensing datasets, namely NWPU-RESISC45, UC-Merced, WHU-RS19, and AID, all of which are exclusively used for testing purposes. Table.~\ref{tab:data} displays the specific dataset partitioning.
\begin{table}[h]
\caption{
DATA COMPOSITION AND SPLIT OF AIFS-DATASET~\cite{li2022rsAIFS}.}
\centering
\begin{tabular}{c|c|c|ccc}
\toprule
Dataset              & Domain & Total & Train & Val & Test \\
\midrule
CUB-200-2011      & \multirow{6}{*}{In}        & 200   & 140   & 30  & 30   \\
Describable Textures &                            & 47    & 33    & 7   & 7    \\
Fungi    &                            & 1394  & 994   & 200 & 200  \\
VGG Flower   &                            & 102   & 71    & 15  & 16   \\
Traffic Signs    &                            & 43    & 30    & 6   & 7    \\
CIFAR100    &                            & 100   & 72    & 13  & 15   \\
\midrule
UCMerced      & \multirow{4}{*}{Out}       & 21    & 0     & 0   & 21   \\
WHU-RS19      &                            & 19    & 0     & 0   & 19   \\
NWPU-RESISC45     &                            & 45    & 0     & 0   & 45   \\
AID      &                            & 30    & 0     & 0   & 30   \\
\bottomrule
\end{tabular}
\label{tab:data}
\end{table}

\textbf{Benchmarking Methods.}
In this article, we compare our method with several SoTA methods for FS-RSSC, such as k-NN~\cite{requeima2019fast_film_cnps}, Finetune~\cite{yosinski2014transferable}, MatchingNet~\cite{vinyals2016matching}, ProtoNet~\cite{snell2017prototypical}, fo-MAML~\cite{finn2017model}, RelationNet~\cite{yang2018relate}, fo-Proto-MAML~\cite{finn2017model}, CNAPs~\cite{requeima2019fast_film_cnps}, SimpleCNAPs~\cite{simcnps}, and DC-DML~\cite{li2022rsAIFS}. These methods have been evaluated on AIFS-DATASET~\cite{li2022rsAIFS}, a large-scale benchmark for FS-RSSC. In addition, we use PMF~\cite{hu2022pmf} as another baseline model, since it has achieved SoTA performance on various FSL benchmarks. We follow the official code and settings of PMF to reproduce its results on the AIFS dataset. For the training settings, all the above methods use the full fine-tuning method. In contrast, our proposed MVP method uses the prompt fine-tuning method.

\textbf{Implementation Details.}
In terms of data organization, we follow the same setup as AIFS-DATASET~\cite{li2022rsAIFS}. Specifically, the number of classes included in each task was randomly selected from the range [5, MAXWAY], while the number of support samples per class was randomly selected from the range [1, MAXSHOT]. The sampling algorithm employed in this process is based on uniform sampling. In the subsequent text, we utilize MW and MS to represent MAXWAY and MAXSHOT, respectively. To comprehensively evaluate model performance in our experiments, we set MW to 5/10/20 and MS to 1/5/10/20. As for the pre-training phase, consistent with PMF~\cite{hu2022pmf}, we also use ViT as the backbone network. ViT is pre-trained on the ImageNet1K dataset using the classical self-supervised mothod DINO~\cite{caron2021emerging_dino} method. In our experiments, we demonstrated results based on ViT-tiny and ViT-small. During the meta-training and meta fine-tuning phases, the parameters of ViT are fixed, and only the newly added prompt parameters are updated.

\subsection{Ablation Study}
In this section, we conduct an ablation study to analyze the effectiveness of the proposed MVP method. First, we compare the performance of meta-visual prompt-tuning and fully fine-tuning and present the results in Table~\ref{tab:token_effect}. Table~\ref{table:params} shows the number of learnable parameters that need to be updated using different fine-tuning methods. Second, we evaluate the effectiveness and computational efficiency of the RPR data augmentation method proposed in this article. Finally, we investigate how the number of prompt tokens affects classification performance and computational efficiency.

\textbf{Is PMF Effective in the Field of Remote Sensing?}
To investigate the performance of the full fine-tuning method in the FS-RSSC task, we conducted an ablation study on AIFS-DATASET using the PMF framework. We compared four settings: 
(1) M1, which is the pre-trained model without any meta-learning process; (2) M2, which is the model after only the meta-training process; (3) M3, which is the model after only the meta-fine-tuning process; (4) M4, which is the complete PMF~\cite{hu2022pmf} model. From the results in Table~\ref{tab:token_effect}, we can draw a conclusion that PMF (full fine-tuning) improves the performance over the pre-trained model without any meta-learning process (comparing models M4, M3, M2 with M1). These suggest that full fine-tuning might not be a good solution for FS-RSSC tasks.

\textbf{Effectiveness of MVP for FS-RSSC and Which Stage to Apply MVP?}
To evaluate the effectiveness of the meta visual prompt-tuning (MVP) method for FS-RSSC, we performed ablation studies under three settings: (1) M5, which only applied MVP for meta-training; (2) M6, which only applied MVP for meta fine-tuning; (3) M9, which applied MVP for both meta-training and meta fine-tuning. The results in Table \ref{tab:token_effect} revealed that: (1) the model with MVP that completes either meta-training or meta fine-tuning process has better performance than the PMF model with the same process (M5 vs M2, and M6 vs M3); (2) Moreover, the model with MVP that completes either the meta-training or meta fine-tuning process even outperforms the PMF model that completes both processes (M5 vs M4 and M6 vs M4); (3) The complete MVP model significantly surpasses the complete PMF model (M9 vs M4). These findings indicate that MVP is an effective and superior method for FS-RSSC, as it can learn from a few examples more efficiently and accurately than PMF.

\begin{table}[t]
\centering
\caption{THE EFFECT OF UPDATING BACKBONE OR PROMPT PARAMETERS ACROSS DIFFERENT META-LEARNING PHASES ON THE AVERAGE CLASSIFICATION ACCURACY (\%) OF AIFS-DATASET UNDER MW5 MS5 AND MW10 MS10 SCENARIOS."-" INDICATES THAT THIS STAGE WILL NOT BE CARRIED OUT.}
\resizebox{\linewidth}{!}{ 
\begin{tabular}{c|c|c|c|c}
\toprule
\multicolumn{3}{c}{Training Configuration} & \multicolumn{2}{c}{Benchmark Results}\\
\midrule
Model  &  Meta Train  & Meta Finetune   & MW5 MS5  & MW10 MS10                       \\
\midrule
M1 & $-$	    & $-$ & 75.5 ± 0.1 & 76.4 ± 0.3 \\ 
M2 & Backbone       & $-$        & 75.6 ± 0.2  & 76.6 ± 0.2    \\
M3 & $-$       & Backbone        & 74.9 ± 0.1  & 77.2 ± 0.3    \\
M4 & Backbone       & Backbone        & 76.6 ± 0.2  & 78.1 ± 0.1    \\
M5 & Prompt       & $-$        & 79.8 ± 0.3  & 80.4 ± 0.2    \\
M6 & $-$       & Prompt        & 76.3 ± 0.2  & 78.6 ± 0.2    \\
M7 & Backbone       & Prompt          & 75.6 ± 0.1  & 78.9 ± 0.2    \\
M8 & Prompt         & Backbone+Prompt & 78.3 ± 0.2  & 80.1 ± 0.4    \\
M9 & Prompt         & Prompt          & 79.6 ± 0.3  & 81.2 ± 0.2    \\
\bottomrule
\end{tabular}
} 
\label{tab:token_effect}
\end{table}

\begin{table}[h] 
\centering
\caption{THE NUMBER OF TRAINABLE PARAMETERS FOR RN18, VIT AND PROMPT VIT WITH 200 PROMPT TOKENS.}
\resizebox{0.85\linewidth}{!}{
\begin{tabular}{c|c|c}
\toprule
Backbone      & Image size & Trainable Params (M)   \\
\midrule
RN18          & 224$\times$224    & 11.28        \\
\midrule
ViT-tiny         & 224$\times$224    & 5.52       \\
ViT-small         & 224$\times$224    & 21.66       \\
ViT-base         & 224$\times$224    & 85.79      \\
\midrule
Prompt ViT-tiny  & 224$\times$224    & 0.46       \\
Prompt ViT-small & 224$\times$224    & 0.92     \\
Prompt ViT-base & 224$\times$224    & 1.84       \\
\bottomrule
\end{tabular}}
\label{table:params}
\end{table}

\begin{figure*}[h] \centering
\subfloat[\label{fig:a}]{
    \includegraphics[scale=0.225]{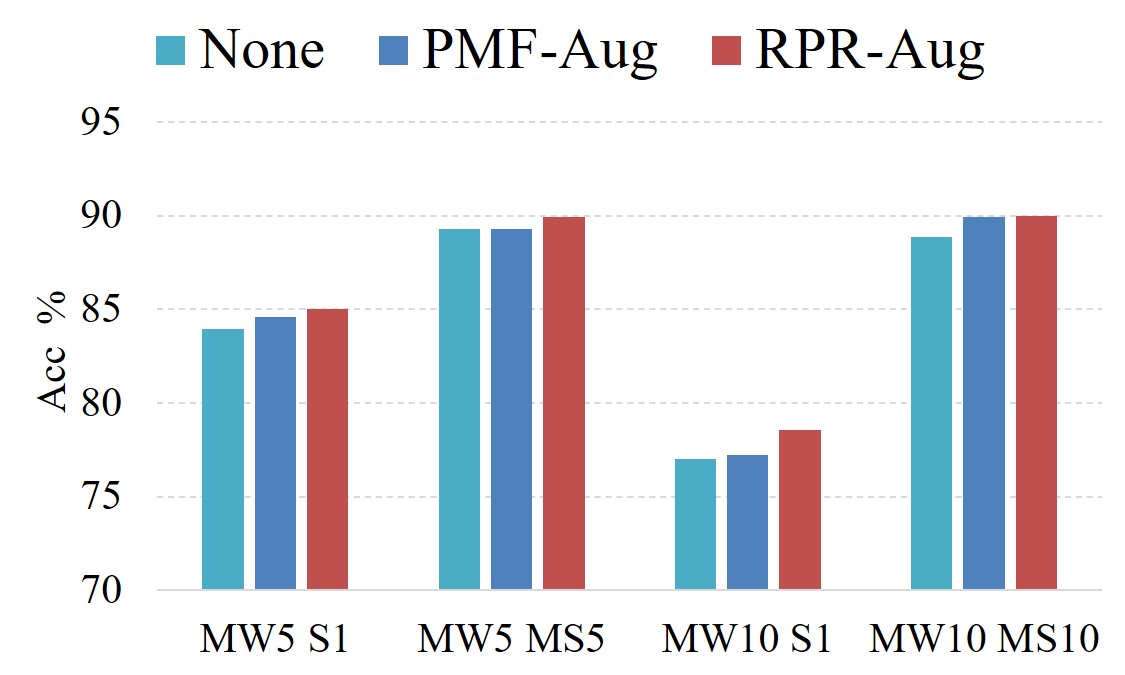}}
\subfloat[\label{fig:b}]{
    \includegraphics[scale=0.225]{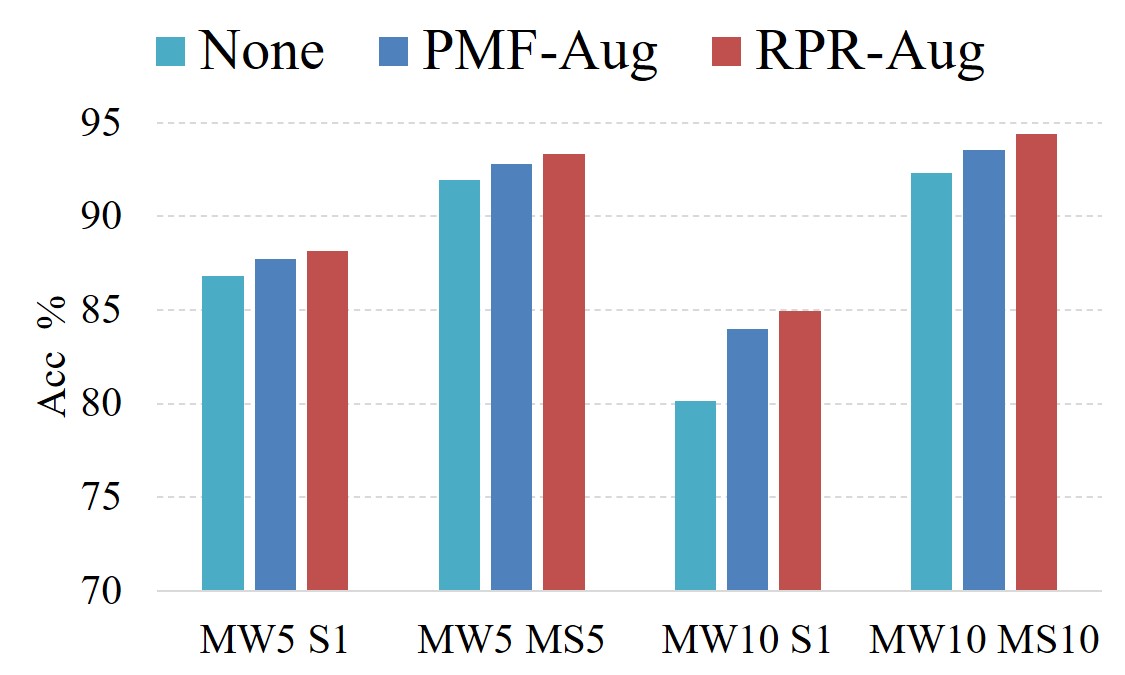}} 
\subfloat[\label{fig:c}]{
    \includegraphics[scale=0.225]{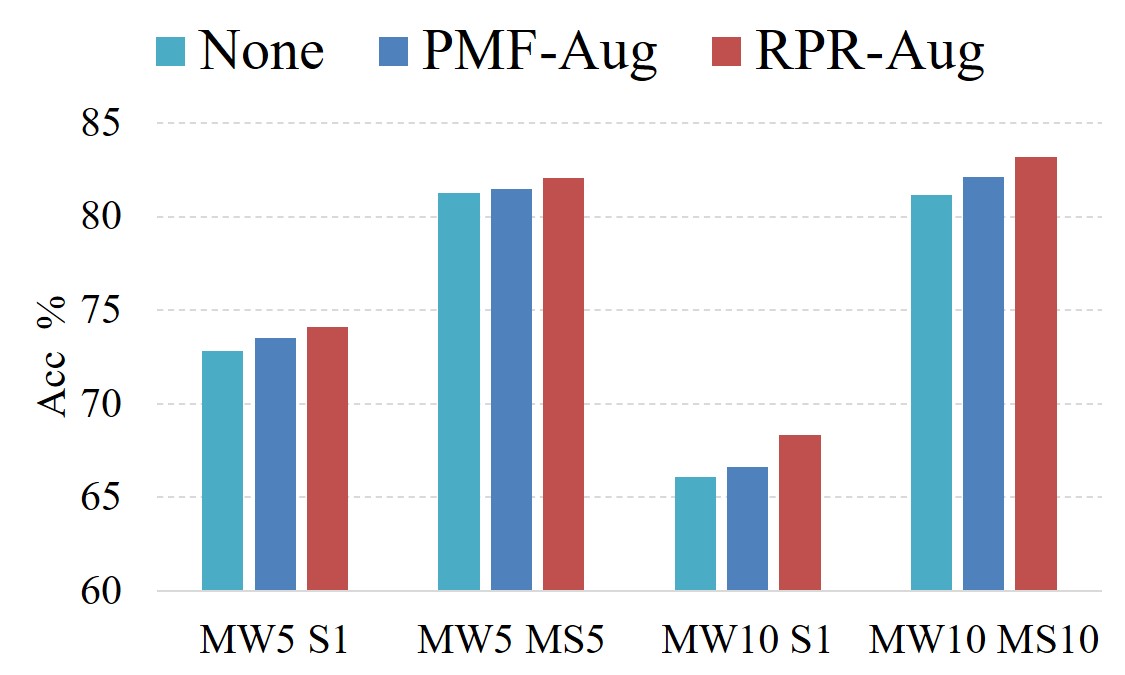}}
\subfloat[\label{fig:d}]{
    \includegraphics[scale=0.225]{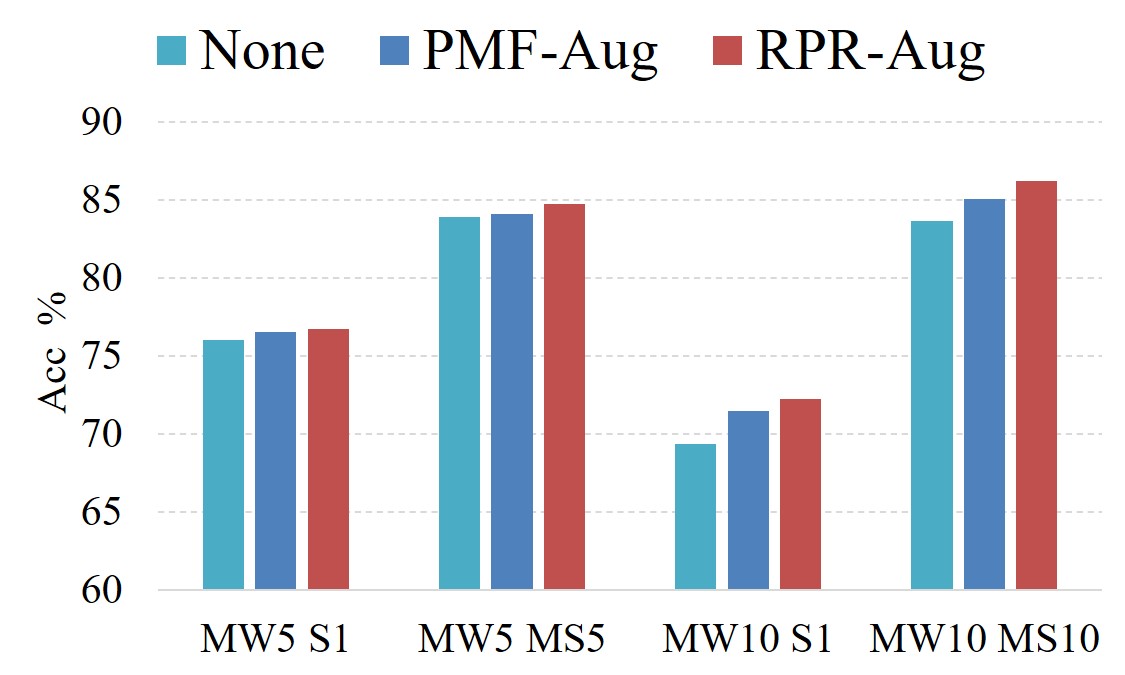} }
\caption{Comparison of different data augmentation methods on various-way-various-shot learning for four out-of-domain datasets. (a) Results on UCM dataset. (b) Results on WHU dataset. (c) Results on NWPU dataset. (d) Results on AID dataset. None denotes no data augmentation, RPR-aug denotes Random Patch Recombination, and PMF-aug denotes the PMF data augmentation method.}
\label{fig:dataaug}
\end{figure*}
\textbf{Combining MVP and Full Fine-Tuning.}
To validate whether MVP and full fine-tuning can work together to achieve better results, we carried out experiments under three settings: (1) M7 applies full fine-tuning in the meta-training stage and MVP tuning in the meta fine-tuning stage; (2) M8 applies MVP tuning in the meta-training stage and full fine-tuning for all parameters in the meta fine-tuning stage; (3) M9 applied MVP tuning for both meta-training and meta fine-tuning. The results in Table \ref{tab:token_effect} revealed that: (1) M7 does not show a significant improvement over M4, indicating that performing MVP only in the meta fine-tuning stage is effective but not remarkable; (2) M8 achieves a substantial improvement over M4, indicating that MVP can help the model obtain a better initialization point which leads to better generalization to new tasks; (3) M9 achieves the SoTA result, demonstrating that employing MVP in both meta-training and meta fine-tuning stages can attain the maximum benefit improvement.

\begin{table}[t]
\caption{RUNNING TIME ANALYSIS ON FS-RSSC TASKS WITH VIT BACKBONE NETWORKS.}
\centering
\resizebox{0.65\linewidth}{!}{ 
\begin{tabular}{c|c|c} 
\toprule
Backbone & Method   & Avg Time (s) \\
\midrule
 \multirow{2}{*}{ViT-tiny} & PMF-Aug   & 3.16                        \\
 &      RPR-Aug                       & \textbf{0.67}               \\
\midrule
\multirow{2}{*}{ViT-small}  & PMF-Aug & 3.08                        \\
 &     RPR-Aug                        & \textbf{0.97}               \\
\midrule
\multirow{2}{*}{ViT-base}  & PMF-Aug      & 3.03                        \\
&      RPR-Aug                        & \textbf{1.61}               \\
\bottomrule 
\end{tabular}
} 
\label{tab:aug_time}
\vspace{-0.3cm} 
\end{table}

\textbf{Effectiveness of RPR Data Augmentation}. This study proposes a novel data augmentation method called Random Patch Recombination (RPR-aug), which is detailed in Sec.~\ref{sec:dataaug}. This section mainly verifies the performance of the RPR method compared to other data augmentation methods. We use the validated PMF~\cite{hu2022pmf} data augmentation method (PMF-aug) as the main comparison method. The PMF data augmentation method includes popular techniques such as mixup, cutmix, color-jitter, translation, and cutout, and activates one or more of these methods based on probability. Similarly to PMF, we apply the proposed RPR to construct pseudo queries based on support images, used as auxiliary tasks in the meta fine-tuning phase. However, unlike PMF, these auxiliary tasks are only used to update prompt parameters while freezing the backbone network. Furthermore, we also automatically selected the learning rate $lr$ and the recombination rate $\alpha$ for each task. We used MVP to choose the optimal $lr$ and $\alpha$ from the ranges $lr \in [1e-4, 1e-3, 1e-2, 0.1, 0]$ and $\alpha \in [0.05, 0.1, 0.2, 0.25]$, and then performed meta fine-tuning with them.

To evaluate the efficacy of RPR, we conducted a series of experiments on four datasets of AIFS (UCM, WHU, NWPU, and AID) for both various-way-various-shot and various-way-one-shot scenarios and compared the results. The outcomes of these experiments are presented in Fig.~\ref{fig:dataaug}. Our findings revealed that overall, RPR outperformed PMF significantly. Specifically, when considering the results across all four tasks (MW5 S1, MW5 MS5, MW10 S1, MW10 MS10), RPR yielded improvements of 0.62\%, 0.7\%, 0.98\%, and 0.68\% compared to PMF on UCM, WHU, NWPU, and AID, respectively. Notably, in 1-shot learning tasks, RPR achieved particularly significant enhancements compared to PMF, with the MW10 S1 task indicating improvements of 1.36\%, 0.97\%, 1.71\%, and 0.74\% on UCM, WHU, NWPU, and AID, respectively. In summary, the proposed RPR data augmentation technique outperforms the PMF method significantly in few-shot image classification tasks, especially in 1-shot learning scenarios.

\begin{table}[h]
\caption{COMPARISON OF PARAMETERS, AVERAGY ACCURACY AND RUNNING TIME FOR DIFFERENT NUMBER OF TOKENS.}
\centering
\resizebox{0.9\linewidth}{!}{ 
\begin{tabular}{c|c|c|c} 
\toprule
Token   & Params  & Avg Acc (\%)   & Time/Iteration (s)  \\
\midrule
10    & 1,150  & 81.9 ± 0.1 & 2.70               \\
20    & 1,210  & 80.3 ± 0.3 & 5.87               \\
50    & 1,385  & 81.8 ± 0.2 & 6.20               \\
200   & 2,542  & 76.5 ± 0.4 & 7.53              \\
\bottomrule
\end{tabular}
} 
\label{tab:token}
\end{table}

\begin{table*}[h]
\caption{IN-DOMAIN AND OUT-OF-DOMAIN ACCURACY OF DIFFERENT MODELS WITH MAXWAY = 5 AND MAXSHOT = 5. `\dag' DENOTES THE RESULTS THAT WE RE-IMPLEMENTED.}
\centering
\resizebox{\linewidth}{!}{ 
\begin{tabular}{l|c|c|cccccc|cccc|c}
\toprule
\multirow{2}{*}{Model} & \multirow{2}{*}{Backbone} & \multirow{2}{*}{Tuning} & \multicolumn{6}{c|}{In-Domain Accuracy (\%)}                                    & \multicolumn{4}{c|}{Out-of-Domain Accuracy (\%)}   & \multirow{2}{*}{Avg}        \\
\cmidrule{4-13}
                       &               &            & CUB        & Textures   & Fungi      & Flower     & Signs          & CIFAR      & UCM        & WHU        & NWPU       & AID        &  \\
\midrule
k-NN~\cite{requeima2019fast_film_cnps}          & RN18         & Full             & 64.1±0.6   & 36.8±0.4   & 45.9±0.6   & 73.8±0.5   & 45.9±0.5      & 50.1±0.5   & 51.8±0.6   & 54.6±0.7   & 42.1±0.5   & 45.9±0.5   & 51.1    \\
Finetune~\cite{yosinski2014transferable}        & RN18         & Full             & 57.0±1.7   & 38.3±1.2   & 45.8±1.8   & 73.9±1.2   & 50.1±1.3      & 54.5±1.4   & 63.6±1.7   & 76.1±1.5   & 55.5±1.6   & 60.2±1.7   & 57.5    \\
MatchingNet~\cite{vinyals2016matching}          & RN18         & Full             & 49.7±0.5   & 36.7±0.4   & 38.2±0.5   & 66.1±0.4   & 52.6±0.4      & 46.9±0.4   & 54.2±0.5   & 65.0±0.5   & 46.9±0.5   & 51.3±0.5   & 50.8   \\
ProtoNet~\cite{snell2017prototypical}           & RN18         & Full             & 44.9±0.5   & 33.5±0.3   & 35.0±0.4   & 56.2±0.5   & 35.7±0.4      & 42.8±0.5   & 51.1±0.5   & 54.2±0.5   & 42.2±0.4   & 44.6±0.4   & 44.0   \\
fo-MAML~\cite{finn2017model}                    & RN18         & Full             & 59.5±0.6   & 38.5±0.4   & 44.0±0.5   & 70.3±0.5   & 49.0±0.4      & 49.7±0.5   & 52.1±0.5   & 60.8±0.5   & 44.5±0.5   & 50.0±0.5   & 51.8   \\
RelationNet~\cite{yang2018relate}               & RN18         & Full             & 60.5±1.9   & 38.8±1.3   & 46.8±1.9   & 72.8±1.4& \textbf{79.9±1.2}& 50.9±1.6   & 56.6±1.8   & 65.4±1.6   & 49.3±1.5   & 52.8±1.6   & 57.4   \\
Proto-MAML~\cite{triantafillou2019metadataset}  & RN18         & Full             & 47.0±1.5   & 33.3±1.1   & 35.5±1.4   & 60.1±1.4   & 36.6±1.2      & 41.9±1.3   & 50.9±1.4   & 52.6±1.4   & 41.6±1.4   & 44.2±1.5   & 44.4   \\
CNAPs~\cite{requeima2019fast_film_cnps}         & RN18         & Full             & 65.6±0.6   & 41.5±0.5   & 46.5±0.5   & 69.7±0.9   & 43.2±0.7      & 55.7±0.5   & 66.9±0.6   & 61.8±0.5   & 49.1±0.5   & 54.6±0.5   & 55.5   \\
SimpleCNAPs~\cite{simcnps}                      & RN18         & Full             & 64.0±0.5   & 44.9±0.9   & 49.8±0.5   & 73.1±0.4   & 48.5±0.4 & \underline{64.8±0.5} & 74.6±0.5   & 74.5±0.5   & 57.0±0.5   & 65.2±0.5   & 61.6   \\
DC-DML~\cite{li2022rsAIFS}                      & RN18         & Full             & 62.2±0.5   & 49.5±0.5   & 50.6±0.6   & 82.2±0.4   & 48.7±0.4      & 59.7±0.5   & 78.9±0.5   & 80.9±0.5   & 60.6±0.5   & 68.6±0.6   & 64.2   \\
PMF-tiny \dag~\cite{hu2022pmf}                  & ViT-t        & Full             & 83.2±0.3   & 61.5±0.2   & 55.6±0.5   & 83.8±0.2   & 43.5±0.2      & 54.6±0.4   & 83.5±0.2   & 88.2±0.1   & 75.3±0.3   & 77.6±0.2   & 70.7   \\
PMF-small \dag~\cite{hu2022pmf}                 & ViT-s        & Full  & 84.3±0.2   & \underline{72.3±0.1}  & 61.8±0.5   & 86.3±0.1   & 53.7±0.5      & 61.3±0.3   & \underline{89.1±0.1}   & \underline{93.0±0.4}   & \underline{80.8±0.2}   & \underline{83.8±0.1}   & \underline{76.6}   \\
\midrule
MVP-tiny (ours)    & ViT-t        & Prompt           & \textbf{87.9±0.3} & 66.1±0.2 & \underline{65.9±0.4}   & \underline{89.1±0.4}   & 55.9±0.3      & 63.5±0.2   & 84.1±0.3   & 89.6±0.4   & 77.2±0.2   & 79.6±0.2   & 75.9   \\
MVP-small (ours)   & ViT-s        & Prompt  & \underline{87.2±0.2}   & \textbf{73.4±0.1} & \textbf{66.9±0.4} & \textbf{91.7±0.3} & \underline{60.3±0.4} & \textbf{67.3±0.2}   & \textbf{89.3±0.3}   & \textbf{93.4±0.1}   & \textbf{81.5±0.2}   & \textbf{84.7±0.2}   & \textbf{79.6}   \\
\bottomrule
\end{tabular}
} 
\label{tab:max5}
\end{table*}

\begin{table*}[h]
\caption{IN-DOMAIN AND OUT-OF-DOMAIN ACCURACY OF DIFFERENT MODELS WITH MAXWAY = 10 AND MAXSHOT = 10. `\dag' DENOTES THE RESULTS THAT WE RE-IMPLEMENTED.}
\centering
\resizebox{\linewidth}{!}{ 
\begin{tabular}{l|c|c|cccccc|cccc|c}
\toprule
\multirow{2}{*}{Model} & \multirow{2}{*}{Backbone} & \multirow{2}{*}{Tuning} & \multicolumn{6}{c|}{In-Domain Accuracy (\%)}                                    & \multicolumn{4}{c|}{Out-of-Domain Accuracy (\%)}   & \multirow{2}{*}{Avg}        \\
\cmidrule{4-13}
                       &               &            & CUB        & Textures   & Fungi      & Flower     & Signs          & CIFAR      & UCM        & WHU        & NWPU       & AID        &  \\
\midrule
k-NN~\cite{requeima2019fast_film_cnps}            & RN18          & Full            & 59.2±1.6   & 36.5±1.2   & 39.3±1.6   & 68.1±1.3   & 42.4±1.3      & 45.0±1.3   & 49.5±1.5   & 52.7±1.5   & 38.2±1.4   & 41.7±1.3   & 47.3   \\
Finetune~\cite{yosinski2014transferable}          & RN18          & Full            & 37.4±1.1   & 31.9±1.2   & 33.7±1.3   & 56.1±1.3   & 46.8±1.1      & 37.6±1.2   & 48.5±1.5   & 61.1±1.3   & 41.8±1.3   & 46.8±1.4   & 44.2   \\
MatchingNet~\cite{vinyals2016matching}            & RN18          & Full            & 42.3±1.5   & 33.5±1.1   & 31.5±1.3   & 58.1±1.4   & 52.9±1.3      & 39.7±1.3   & 48.9±1.5   & 58.3±1.3   & 40.0±1.4   & 46.0±1.3   & 45.1   \\
ProtoNet~\cite{snell2017prototypical}             & RN18          & Full            & 35.4±1.2   & 29.1±1.1   & 24.2±1.1   & 42.1±1.3   & 28.4±1.0      & 34.8±1.2   & 42.3±1.3   & 45.9±1.3   & 33.9±1.3   & 35.1±1.2   & 35.1   \\
fo-MAML~\cite{finn2017model}                      & RN18          & Full            & 51.3±1.5   & 35.6±1.3   & 37.5±1.4   & 61.5±1.3   & 43.2±1.2      & 40.3±1.3   & 50.1±1.4   & 58.2±1.4   & 39.2±1.4   & 44.8±1.3   & 46.2   \\
RelationNet~\cite{yang2018relate}                 & RN18          & Full            & 44.9±1.5   & 33.1±1.2   & 35.7±1.7   & 58.3±1.5   & \textbf{73.4±1.3}      & 36.8±1.4   & 49.1±1.5   & 59.8±1.3   & 40.2±1.4   & 46.0±1.5   & 47.7   \\
Proto-MAML~\cite{triantafillou2019metadataset}    & RN18          & Full            & 37.5±1.4   & 28.3±1.0   & 27.0±1.0   & 48.3±1.4   & 28.4±1.0      & 33.0±1.2   & 39.3±1.4   & 41.6±1.5   & 32.2±1.1   & 33.6±1.3   & 34.9   \\
CNAPs~\cite{requeima2019fast_film_cnps}           & RN18          & Full            & 59.5±0.5   & 44.4±0.4   & 46.6±0.5   & 70.1±0.4   & 55.8±0.5      & 52.8±0.5   & 61.7±0.5   & 60.6±0.5   & 44.3±0.5   & 52.0±0.5   & 54.8   \\
SimpleCNAPs~\cite{simcnps}                        & RN18         & Full             & 64.6±0.5   & 40.4±0.4   & 47.8±0.6   & 68.6±0.4   & 54.3±0.5      & 51.6±0.5   & 62.1±0.5   & 61.5±0.5   & 43.6±0.5   & 52.7±0.5   & 54.7   \\
DC-DML~\cite{li2022rsAIFS}                        & RN18         & Full             & 56.3±0.5   & 45.7±0.5   & 43.1±0.6   & 71.9±0.4   & 47.5±0.4      & 53.7±0.5   & 69.4±0.6   & 72.2±0.5   & 49,7±0.6   & 56.8±0.6   & 57.4    \\
PMF-tiny \dag~\cite{hu2022pmf}                    & ViT-t        & Full             & 85.9±0.1   & 67.6±0.4   & 54.4±0.4   & \underline{89.1±0.1}   & 51.5±0.4      & 60.2±0.6   & 86.2±0.2   & 90.4±0.5   & 78.2±0.3   & 80.5±0.3   & 74.4   \\
PMF-small \dag~\cite{hu2022pmf}                   & ViT-s        & Full             & 86.1±0.1   & \underline{75.5±0.1}   & 60.4±0.4   & 87.2±0.1   & 58.3±0.3      & \underline{62.7±0.4}   & \underline{89.8±0.3}   & \underline{93.8±0.2}   & \underline{81.8±0.1}   & \underline{84.8±0.1}   & \underline{78.1}   \\
\midrule
MVP-tiny (ours)                                   & ViT-t        & Prompt           & \textbf{89.1±0.3}   & 68.9±0.1   & \underline{66.0±0.3}   & 89.0±0.2   & 56.9±0.3      & 64.4±0.4   & 86.8±0.1   & 90.0±0.4   & 78.6±0.3   & 82.1±0.2   & 77.3   \\
MVP-small (ours)                                  & ViT-s        & Prompt           & \underline{88.8±0.2}   & \textbf{76.2±0.2}   & \textbf{67.4±0.1}   & \textbf{92.4±0.3}   & \underline{64.9±0.4}      & \textbf{69.3±0.3}   & \textbf{89.9±0.2}   & \textbf{94.4±0.3}   & \textbf{83.2±0.4}   & \textbf{85.1±0.1}   & \textbf{81.2}   \\
\bottomrule 
\end{tabular}
}
\label{tab:max10}
\end{table*}

\begin{table*}[h]
\caption{IN-DOMAIN AND OUT-OF-DOMAIN ACCURACY OF DIFFERENT MODELS WITH MAXWAY = 20 AND MAXSHOT = 20. `\dag' DENOTES THE RESULTS THAT WE RE-IMPLEMENTED.}
\centering
\resizebox{\linewidth}{!}{ 
\begin{tabular}{l|c|c|cccccc|cccc|c}
\toprule
\multirow{2}{*}{Model} & \multirow{2}{*}{Backbone} & \multirow{2}{*}{Tuning} & \multicolumn{6}{c|}{In-Domain Accuracy (\%)}                                    & \multicolumn{4}{c|}{Out-of-Domain Accuracy (\%)}   & \multirow{2}{*}{Avg}        \\
\cmidrule{4-13}
                       &               &            & CUB        & Textures   & Fungi      & Flower     & Signs          & CIFAR      & UCM        & WHU        & NWPU       & AID        &  \\
\midrule
k-NN~\cite{requeima2019fast_film_cnps}        & RN18  & Full   & 49.7±0.6 & 45.4±0.4 & 32.7±0.5 & 72.5±0.4 & 48.5±0.5 & 44.9±0.5 & 49.2±0.5 & 53.8±0.5 & 35.7±0.5 & 40.5±0.5  & 47.3  \\
Finetune~\cite{yosinski2014transferable}      & RN18  & Full   & 33.3±1.0 & 40.4±0.7 & 27.5±0.7 & 54.6±1.0 & 47.8±1.0 & 37.0±1.0 & 46.9±1.0 & 56.1±1.0 & 37.1±1.0 & 41.4±1.0  & 42.2  \\
MatchingNet~\cite{vinyals2016matching}        & RN18  & Full   & 38.5±0.6 & 41.8±0.4 & 27.4±0.4 & 63.9±0.5 & 58.8±0.5 & 40.5±0.5 & 44.7±0.6 & 54.6±0.5 & 35.9±0.5 & 40.6±0.5  & 44.7  \\
ProtoNet~\cite{snell2017prototypical}         & RN18  & Full   & 30.8±0.6 & 40.0±0.4 & 20.1±0.4 & 45.2±0.6 & 31.5±0.4 & 32.8±0.6 & 38.2±0.6 & 41.7±0.7 & 29.4±0.6 & 30.5±0.6  & 34.1  \\
fo-MAML~\cite{finn2017model}                  & RN18  & Full   & 45.2±0.6 & 42.9±0.4 & 31.4±0.5 & 63.7±0.5 & 48.1±0.5 & 39.5±0.6 & 42.4±0.6 & 48.8±0.6 & 32.4±0.5 & 36.3±0.5  & 43.1  \\
RelationNet~\cite{yang2018relate}             & RN18  & Full   & 35.3±0.6 & 38.0±0.4 & 27.6±0.5 & 65.5±0.5 & 85.7±0.3 & 34.3±0.5 & 40.9±0.6 & 50.7±0.6 & 31.9±0.5 & 37.5±0.6  & 44.7  \\
Proto-MAML~\cite{triantafillou2019metadataset}& RN18  & Full   & 35.4±1.1 & 39.4±0.7 & 22.7±0.7 & 49.8±1.1 & 33.8±1.0 & 34.0±1.0 & 36.4±1.1 & 40.5±1.1 & 28.4±1.0 & 30.7±1.0  & 35.1  \\
CNAPs~\cite{requeima2019fast_film_cnps}       & RN18  & Full   & 54.2±0.6 & 48.9±0.4 & 41.5±0.6 & 77.0±0.4 & 62.7±0.5 & 51.2±0.6 & 57.8±0.5 & 52.5±0.6 & 38.8±0.6 & 44.2±0.6  & 52.9  \\
SimpleCNAPs~\cite{simcnps}                    & RN18  & Full   & 54.9±0.6 & 47.8±0.4 & 41.7±0.6 & 74.2±0.4 & 71.1±0.4 & 48.0±0.6 & 55.6±0.6 & 50.3±0.7 & 40.0±0.6 & 45.5±0.6  & 52.9  \\
DC-DML~\cite{li2022rsAIFS}                    & RN18  & Full   & 48.6±0.7 & 54.5±0.4 & 36.0±0.6 & 76.9±0.4 & 51.3±0.4 & 53.0±0.6 & 62.8±0.6 & 63.0±0.6 & 43.3±0.7 & 48.6±0.7  & 53.8  \\
PMF-tiny \dag~\cite{hu2022pmf}                & ViT-t & Full   & 84.4±0.2 & 69.1±0.3 & 49.5±0.6 & 84.5±0.6 & 40.8±0.4 & 52.7±0.5 & 81.3±0.2 & 86.2±0.1 & 70.3±0.4 & 74.9±0.4  & 69.4 \\
PMF-small \dag~\cite{hu2022pmf}               & ViT-s & Full   & 84.1±0.3 & \underline{80.3±0.4} & 55.7±0.2 & 85.6±0.2 & \underline{74.7±0.3} & 60.4±1.5 & \underline{88.2±0.3} & \underline{93.5±0.1} & \underline{77.8±0.2} & \underline{83.0±1.1}  & \underline{78.3} \\
\midrule
MVP-tiny (ours)                               & ViT-t & Prompt & \underline{87.2±0.3} & 72.2±0.3 & \underline{61.3±0.4} & \underline{89.9±0.4} & 64.4±0.5 & \underline{63.4±0.5} & 85.6±0.2 & 90.1±0.1 & 76.9±0.3 & 79.9±0.3  & 77.1 \\
MVP-small (ours)                              & ViT-s & Prompt & \textbf{88.9±0.2} & \textbf{80.9±1.4} & \textbf{67.1±0.3} & \textbf{94.3±0.6} & \textbf{78.7±0.5} & \textbf{67.1±0.4} & \textbf{89.7±0.1} & \textbf{93.7±0.2} & \textbf{81.8±0.4} & \textbf{84.3±0.2}  & \textbf{82.1} \\
\bottomrule
\end{tabular}
} 
\label{tab:max20}
\end{table*}


\textbf{Efficiency of RPR Data Augmentation.} 
To compare the efficiency of our RPR-aug method and the PMF-aug method, we further conducted experiments on AIFS using different backbone networks of ViT-tiny and ViT-small. We tested six 10-way $k$-shot tasks, where $k \in [1,2,4,6,8,10]$, and repeated 1000 data augmentation experiments for each task. To ensure the fairness of the comparison, we set the recombination rate $\alpha$ of RPR-aug to a maximum value of 0.25 and used the same model and same batch. Table~\ref{tab:aug_time} shows the total average running time of 1000 experiments for each task. From Table~\ref{tab:aug_time}, we can see that RPR-aug is 3 times faster than PMF-aug on average, indicating that our RPR-aug method is more efficient and more suitable for FS-RSSC tasks based on ViT.

\textbf{Prompt Tokens Number.} 
This is an important hyper-parameter needed to tune
for MVP, we carried experiment to test the effect of the number of prompt tokens on the performance and efficiency of the model. To conduct comparative experiments, we designed our experimental setup following the principle of controlling variables. We used the NWPU dataset of AIFS as a benchmark and evaluated the performance of the MVP model on four RS-FSSC tasks, including 5-way 1-shot, 5-way 5-shot, 10-way 1-shot, and 10-way 10-shot. Table~\ref{tab:token} shows a comparison of the performance of the MVP model with different numbers of prompt tokens loaded in these four tasks. As can be seen from the figures, when the number of tokens is set to 10, the MVP model achieves the highest classification accuracy in most tasks. Moreover, we observed that as the number of tokens increases, so does the time consumption of the model. In summary, considering the trade-off between performance accuracy and computational efficiency, we fixed the number of prompt tokens for the MVP model at 10 in this article and applied this setting consistently across all tasks.

\subsection{Comparison With SoTA Methods}
\textbf{Main Results and Comparisons.}
We present the experimental results obtained under three distinct evaluation benchmarks, where MW and MS were set to 5, 10, and 20, respectively, in Tables~\ref{tab:max5}-\ref{tab:max20}. Comparing these results, we can deduce that MVP outperforms the SoTA methods in terms of average accuracy. Specifically, our MVP method achieves significant improvements over the strong baseline PMF method, with average increases of 3.0\%, 3.1\%, and 3.8\% at MW and MS of 5/10/20. These findings support our view that prompt-tuning techniques are better suited for few-shot classification tasks than full-tuning techniques.

In the context of in-domain tasks, MVP exhibits considerable advantages over other SoTA methods, particularly in fine-grained evaluation benchmarks such as CUB, Fungi, and Flower datasets. In these datasets, MVP surpasses all other listed methods in terms of performance. This observation further highlights the superior capabilities of the MVP technology in fine-grained few-shot classification tasks. Notably, when compared with the PMF method, the MVP approach demonstrates the most prominent advantages in the Fungi dataset at MW and MS of 5/10/20, with improvements of 5.1\%, 7\%, and 11.4\%, respectively.

Regarding out-of-domain tasks, our analysis indicates that MVP generally outperforms the second-best PMF method. This trend is particularly noticeable in the NWPU dataset, where MVP improves by 0.7\%, 1.4\%, and 4\% at MW and MS of 5/10/20. Moreover, These results suggest that MVP has better generalization ability when dealing with new unseen FS-RSSC tasks. 

\textbf{Analysis of Model Parameters and Performance.}
By utilizing Tables~\ref{tab:max5}-\ref{tab:max20} and Table~\ref{table:params}, we can perform a comprehensive analysis of how model parameters impact the performance and accuracy of our model. Our study demonstrates that, with the use of a meta-based prompt-tuning framework, ViT-tiny achieves a significant improvement in performance compared to RN18 despite having similar parameter counts. Specifically, in the out-of-domain FS-RSSC task, our MVP-tiny method shows an average accuracy increase of 19.2\% over the DC-DML~\cite{li2022rsAIFS} method based on RN18 across all three evaluation benchmarks. Furthermore, while ViT-small possesses four times as many parameters as ViT-tiny, it does not exhibit a significant increase in accuracy for the FS-RSSC task. Nonetheless, the MVP-small method only shows an average precision increase of 3.9\% compared to MVP-tiny across all three evaluation benchmarks. Therefore, our results suggest that deploying applications based on MVP-tiny can provide a more balanced trade-off between efficiency and performance.

\textbf{The Impact of Domain Shift on Classification Accuracy.} 
The cross-domain results shown in Table~\ref{tab:max10} and Table~\ref{tab:max20} indicated that all algorithms suffered from a varying degree of accuracy degradation when dealing with the MW20 MS20 task compared to the MW10 MS10 task. This phenomenon was absent in the in-domain scenarios. This suggested that domain shift, especially when the number of categories increased, had a significant effect on classification accuracy. However, the results also revealed that the MVP model was more robust to domain shift. In particular, the MVP achieved high accuracy on out-of-domain datasets while effectively reducing the effect of domain shift. On the UCM, WHU, NWPU, and AID datasets, MVP-small and MVP-tiny showed an average accuracy drop of 1.34\% and 1.5\%, respectively, while PMF-small and PMF-tiny showed an average drop of 1.95\% and 5.65\%, and DC-DML showed an average drop of 7.6\%. Hence, the MVP model outperformed other models in cross-domain FS-RSSC tasks, especially when the number of categories was large.

\begin{figure}[t] \centering
\centering
\subfloat[\label{fig:a}]{
    \includegraphics[scale=0.34]{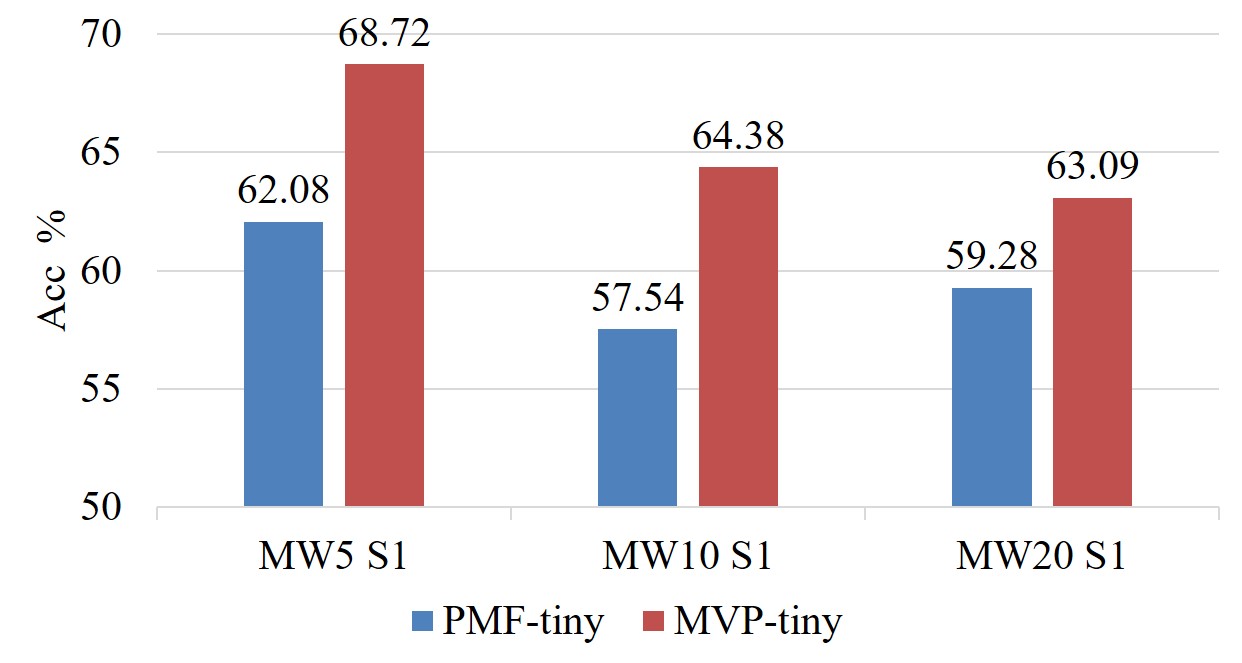}} \\
\subfloat[\label{fig:b}]{
    \includegraphics[scale=0.34]{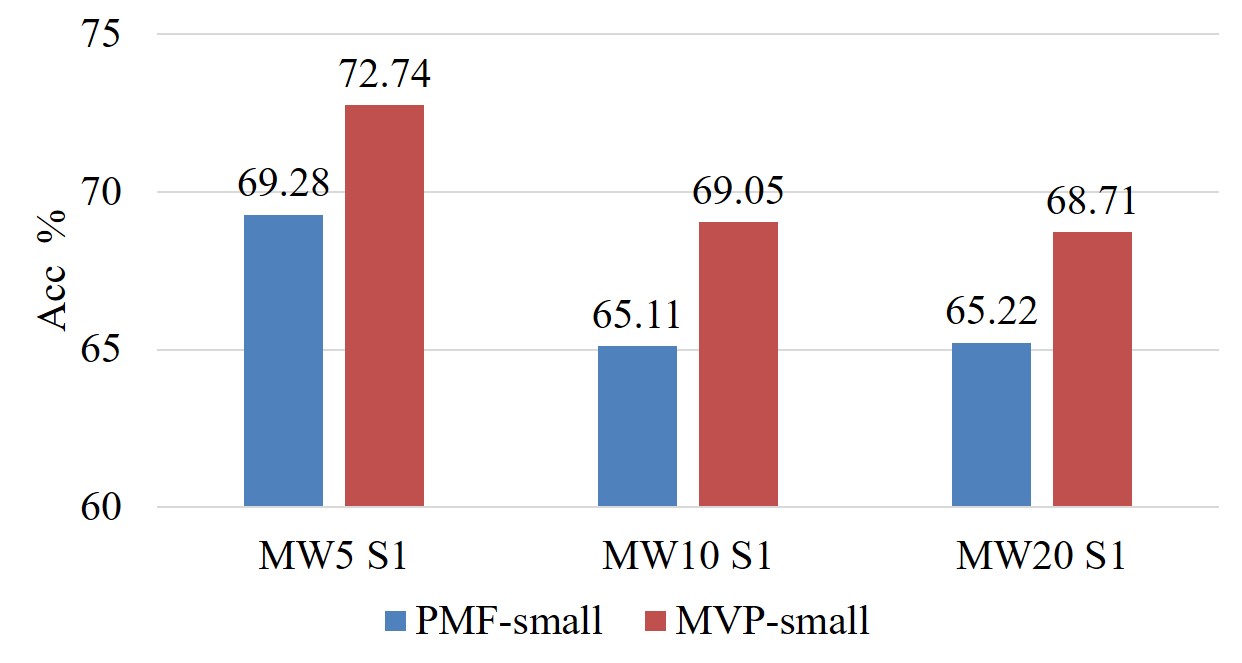}}
\caption{Results of the PMF~\cite{hu2022pmf} and our MVP model on various-way-one-shot learning. (a) Results with ViT-tiny backbone network. (b) Results with ViT-small backbone network.}  
\label{fig:1shot}
\end{figure}

\textbf{Results for One-Shot Learning.}
Fig.~\ref{fig:1shot} presents the results of our model on the one-shot learning task with one support image per category. Our MVP method outperforms the strong baseline PMF method in all one-shot Learning tasks. Specifically, when using ViT-tiny as the backbone network, MVP-tiny achieves better results than PMF-tiny in all ten evaluation tasks. Averaging all evaluation tasks, MVP-tiny improves by 6.64\%, 6.84\%, and 3.80\% compared to PMF-tiny when MW and MS are 5/10/20 respectively; MVP-small improves by 3.46\%, 3.94\%, and 3.49\% compared to PMF-small when MW and MS are 5/10/20 respectively. Averaging all out-of-domain tasks, MVP-tiny improves by 2.44\%, 3.07\%, and 2.73\% compared to PMF-tiny in these three tasks respectively; MVP-small improves by 0.82\%, 0.62\%, and 0.12\% compared to PMF-small in these three tasks respectively. These results demonstrate that our MVP method has better generalization ability in the challenging 1-shot learning task.

\begin{figure}[t] 
\centering
\subfloat[\label{}]{
    \includegraphics[scale=0.25]{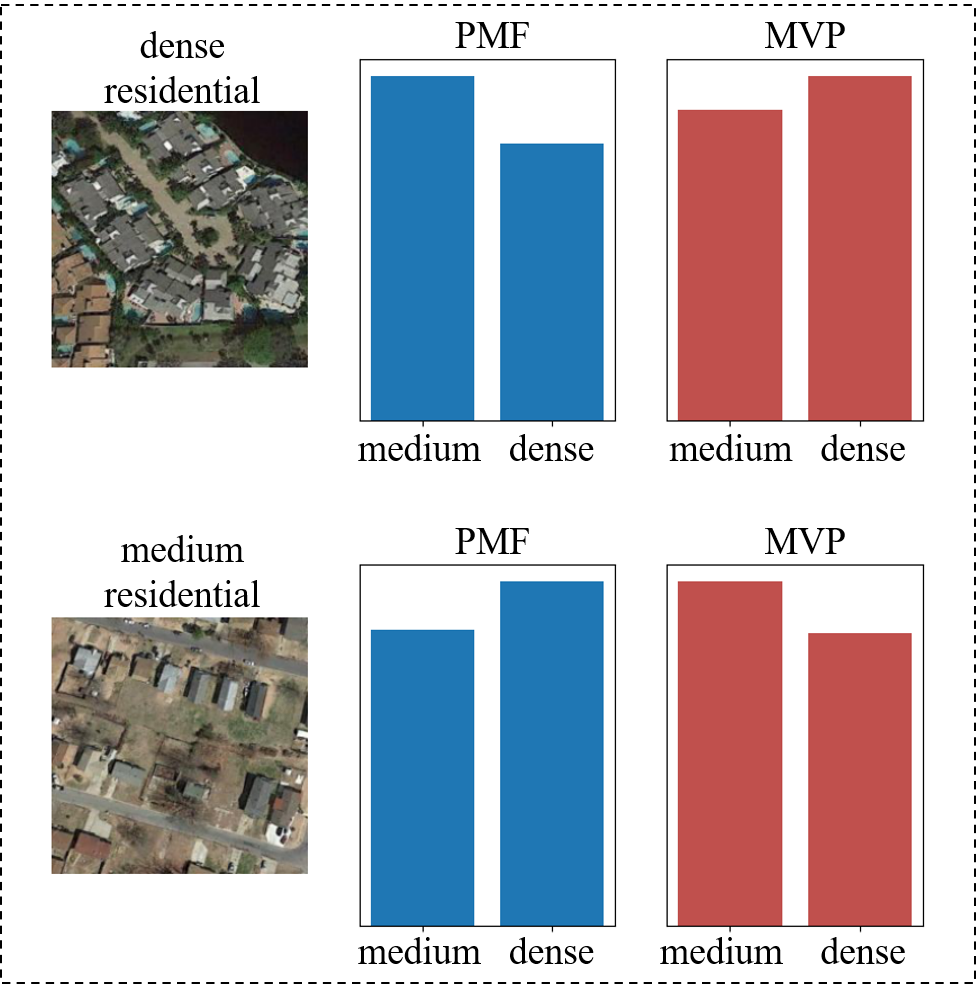}}
\subfloat[\label{}]{
    \includegraphics[scale=0.25]{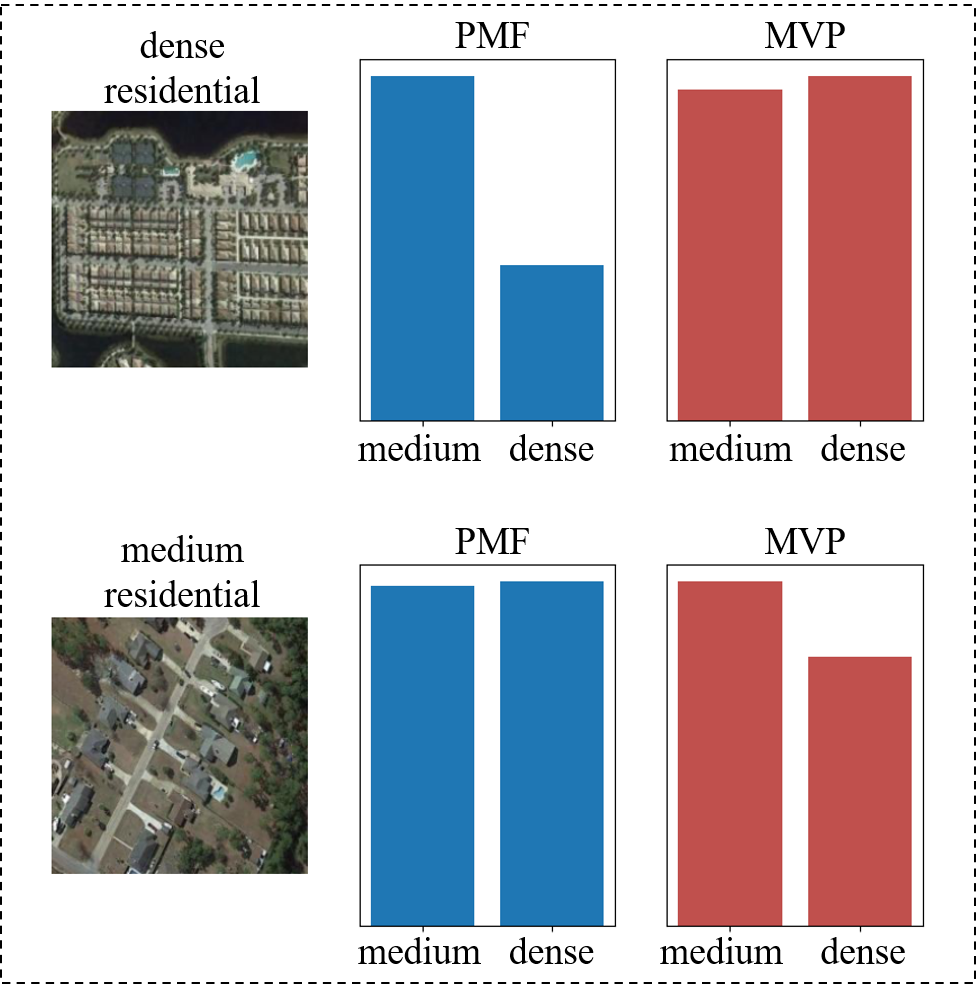}} \\
\subfloat[\label{}]{
    \includegraphics[scale=0.25]{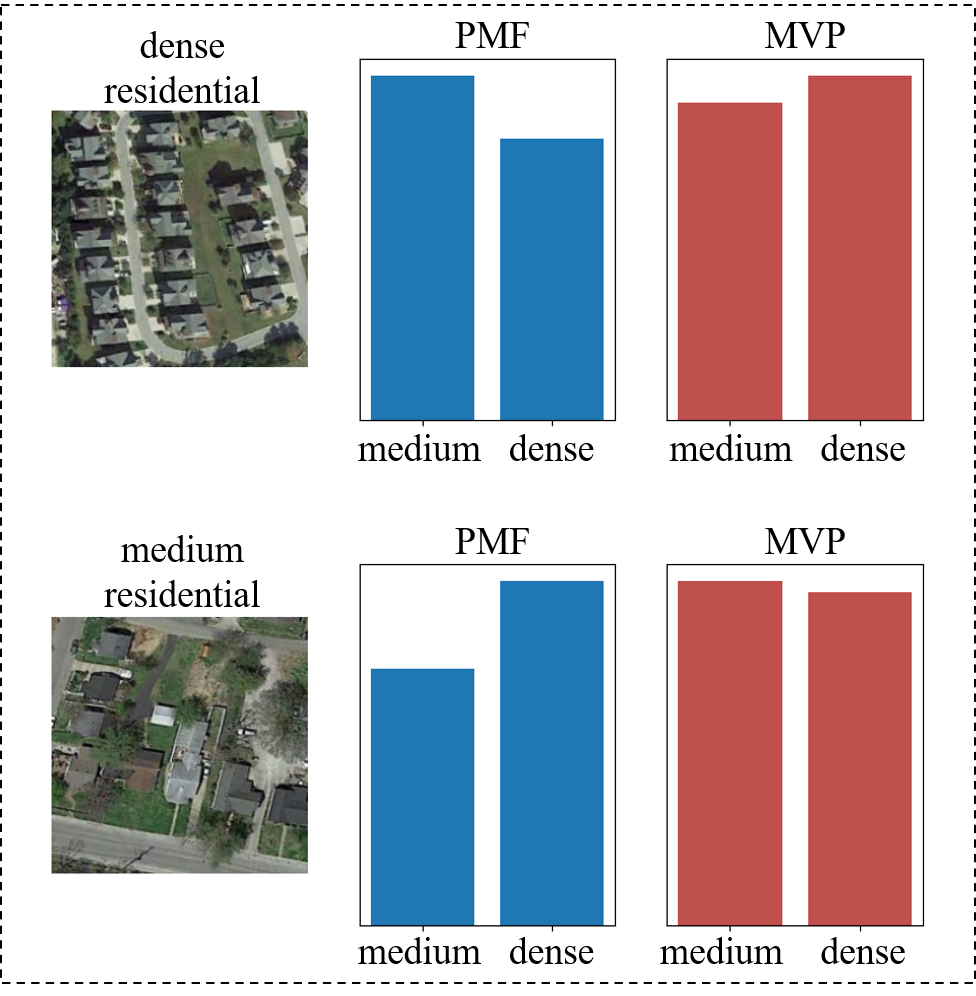}}
\subfloat[\label{}]{
    \includegraphics[scale=0.25]{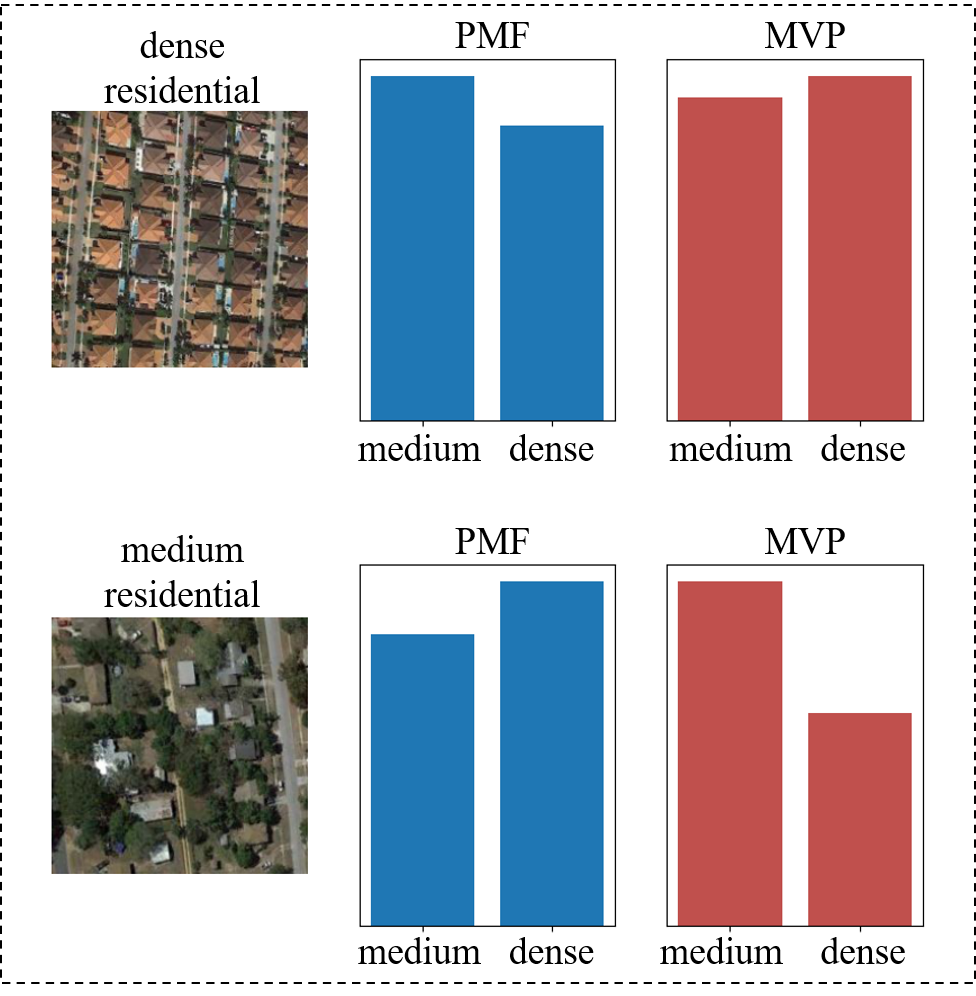} }
    \caption{Qualitative results of few-shot classification for PMF and ours MVP.}
    \label{fig:qa}
\end{figure}

\textbf{Qualitative Analysis of Few-Shot Classification.}
The aim of this section is to examine the disparities between the outputs generated by the MVP and PMF models. Based on the in-domain results presented in Tables~\ref{tab:max5} to \ref{tab:max20}, it was observed that the MVP model demonstrated a notable advantage in processing fine-grained classification tasks. This led us to hypothesize that the transfer of fine-grained classification abilities may contribute to the superior performance of the MVP model in downstream tasks. To evaluate this hypothesis, an experiment was conducted in which a 10-way 5-shot task was extracted from the NWPU dataset, distinct from the AIFS-in-domain dataset utilized for training the models. The outputs generated by each model were subsequently compared for every query image. A meticulous examination was performed on categories accurately predicted by the MVP model but mispredicted by the PMF model. Notably, when comparing two analogous categories: ``dense residential" and ``medium residential", it was determined that the MVP model achieved a significantly higher average accuracy than the PMF model. To further scrutinize this phenomenon, a 2-way 1-shot experiment was designed. As depicted in Fig.~\ref{fig:qa}, it is apparent that the MVP model is better equipped to handle fine-grained classification problems. In summary, these observations indicate that the MVP model possesses the ability to learn more discriminative and transferable features, thereby facilitating the effective transfer of fine-grained classification abilities to downstream tasks.

\section{Conclusion}
In this article, we focused on the challenging and practical scenario of few-shot remote sensing scene classification (FS-RSSC), where the goal is to classify remote sensing images into different categories using only a few labeled examples for each category. 
To tackle this problem, we proposed a novel and efficient method called Meta Visual Prompt Tuning (MVP) that leverages prompt tuning and meta-learning to adapt a pre-trained visual transformer (ViT) model to new tasks with minimal data and resources. 
Specifically, MVP has three main components: 
(1) prompt tuning, a parameter-efficient fine-tuning technique that updates only the newly added prompt parameters and freezes the rest of the model, which reduces the storage demand and mitigates the overfitting risk; 
(2) meta-learning, a fast adaptation technique that learns to initialize the prompt parameters from multiple source tasks and rapidly adapts them to new tasks with only a few gradient steps, which facilitates cross-domain adaptation; 
and (3) data augmentation, a novel technique that operates on the patch embeddings of the input tokens of transformer blocks to enhance the scene representation and diversity. 
We evaluated MVP on a realistic cross-domain FS-RSSC benchmark dataset and demonstrated its superior performance over existing methods. MVP achieved especially remarkable results on vary-way, vary shot, and one-shot tasks, which are more challenging and relevant for real-world applications. 
Our work paved the way for applying ViT models to FS-RSSC tasks and provided a solution that is suitable for the deployment platform of FS-RSSC algorithms. 
In the future, we expect to extend the training data domain to include more data distributions, which may further improve the accuracy and robustness of prompt tuning-based methods for FS-RSSC tasks, especially for enhancing cross-domain performance.

\appendices

\ifCLASSOPTIONcaptionsoff
  \newpage
\fi

\bibliographystyle{IEEEtran}

\bibliography{IEEEabrv,IEEEtran}

\end{document}